\documentclass[sigconf, nonacm]{acmart}
\usepackage{xspace}
\usepackage[ruled,vlined,linesnumbered]{algorithm2e}
\usepackage{threeparttable}
\usepackage{multirow}
\usepackage{array}
\usepackage{footnote}
\usepackage{colortbl}
\usepackage{url}
\usepackage[hypcap=true]{caption}

\usepackage{booktabs,colortbl,xltabular,ragged2e}
\definecolor{layergray}{RGB}{219,219,219} % 层次区分浅灰

\renewcommand\footnotetextcopyrightpermission[1]{} % removes footnote with conference information in first column

\newcommand{\boldres}[1]{{\textbf{\textcolor[HTML]{C82423}{#1}}}}
\newcommand{\secondres}[1]{{\underline{\textcolor{blue}{#1}}}}

\newcommand{\model}{\texttt{E$^3$Former}\xspace}
\newcommand{\modelEGD}{\texttt{E$^3$Former-OS}\xspace}
\newcommand{\modelFTPL}{\texttt{E$^3$Former-FTPL}\xspace}

\newtheorem{myDef}{Definition}

\newtheorem{myTheo}{Theorem}

\definecolor{light-gray}{gray}{0.89}

\newcommand\vldbdoi{XX.XX/XXX.XX}

\begin{document}
% \title{Online Ensemble Transformer for Cloud Resource Forecasting in Predictive Auto-Scaling}
\title{Online Ensemble Transformer for Accurate Cloud Workload Forecasting in Predictive Auto-Scaling}

%%
%% The "author" command and its associated commands are used to define the authors and their affiliations.
\author{Jiadong Chen}
\affiliation{%
  \institution{Shanghai Jiao Tong Univ.}
  \city{Shanghai}
  \state{China}
}

\author{Xiao He}
\affiliation{%
   \institution{ByteDance Inc.}
  \city{Hangzhou}
  \country{China}
}
% \email{xiao.hx@bytedance.com}

\author{Hengyu Ye}
\affiliation{%
  \institution{Shanghai Jiao Tong Univ.}
  \city{Shanghai}
  \state{China}
}
% \email{cs_22_yhy@sjtu.edu.cn}

% \email{chenjiadong998@sjtu.edu.cn}

\author{Fuxin Jiang}
\affiliation{%
  \institution{ByteDance Inc.}
  \city{Beijing}
  \country{China}
}
% \email{jiangfuxin@bytedance.com}

\author{Tieying Zhang}
\affiliation{%
  \institution{ByteDance Inc.}
  \city{San Jose}
  \country{USA}
}
% \email{tieying.zhang@bytedance.com}

\author{Jianjun Chen}
\affiliation{%
  \institution{ByteDance Inc.}
  \city{San Jose}
  \country{USA}
}
% \email{jianjun.chen@bytedance.com}

\author{Xiaofeng Gao}
\affiliation{%
  \institution{Shanghai Jiao Tong Univ.}
  \city{Shanghai}
  \state{China}
}
% \email{gao-xf@cs.sjtu.edu.cn}

\begin{abstract}
In the swiftly evolving domain of cloud computing, the advent of serverless systems underscores the crucial need for predictive auto-scaling systems. 
This necessity arises to ensure optimal resource allocation and maintain operational efficiency in inherently volatile environments.
At the core of a predictive auto-scaling system is the workload forecasting model.
Existing forecasting models struggle to quickly adapt to the dynamics in online workload streams and have difficulty capturing the complex periodicity brought by fine-grained, high-frequency forecasting tasks.
Addressing this, we propose a novel online ensemble model, \model, for online workload forecasting in large-scale predictive auto-scaling. 
Our model synergizes the predictive capabilities of multiple subnetworks to surmount the limitations of single-model approaches, thus ensuring superior accuracy and robustness. 
Remarkably, it accomplishes this with a minimal increase in computational overhead, adhering to the lean operational ethos of serverless systems. 
Through extensive experimentation on real-world workload datasets, we establish the efficacy of our ensemble model. In online forecasting tasks, the proposed method reduces forecast error by an average of 10\%, and its effectiveness is further demonstrated through a predictive auto-scaling test in the real-life online system.
Currently, our method has been deployed within ByteDance's Intelligent Horizontal Pod Auto-scaling (IHPA) platform, which supports the stable operation of over 30 applications, such as Douyin E-Comerce, TouTiao, and Volcano Engine. The predictive auto-scaling capacity reaching over 600,000 CPU cores. On the basis of essentially ensuring service quality, the predictive auto-scaling system can reduce resource utilization by over 40\%.

\end{abstract}

\maketitle

% \pagestyle{\vldbpagestyle}
% \begingroup\small\noindent\raggedright\textbf{PVLDB Reference Format:}\\
% \vldbauthors. \vldbtitle. PVLDB, \vldbvolume(\vldbissue): \vldbpages, \vldbyear.\\
% \href{https://doi.org/\vldbdoi}{doi:\vldbdoi}
% \endgroup
% \begingroup
% \renewcommand\thefootnote{}\footnote{\noindent
% This work is licensed under the Creative Commons BY-NC-ND 4.0 International License. Visit \url{https://creativecommons.org/licenses/by-nc-nd/4.0/} to view a copy of this license. For any use beyond those covered by this license, obtain permission by emailing \href{mailto:info@vldb.org}{info@vldb.org}. Copyright is held by the owner/author(s). Publication rights licensed to the VLDB Endowment. \\
% \raggedright Proceedings of the VLDB Endowment, Vol. \vldbvolume, No. \vldbissue\ %
% ISSN 2150-8097. \\
% \href{https://doi.org/\vldbdoi}{doi:\vldbdoi} \\
% }\addtocounter{footnote}{-1}\endgroup

% \ifdefempty{\vldbavailabilityurl}{}{
% \vspace{.3cm}
% \begingroup\small\noindent\raggedright\textbf{PVLDB Artifact Availability:}\\
% The source code, data, and/or other artifacts have been made available at \url{https://github.com/WI114rD/E3Former}.
% \endgroup
\section{Introduction}

\begin{figure}[htbp]
% \vspace{-3mm}
\centering
\includegraphics[width=0.95\linewidth]{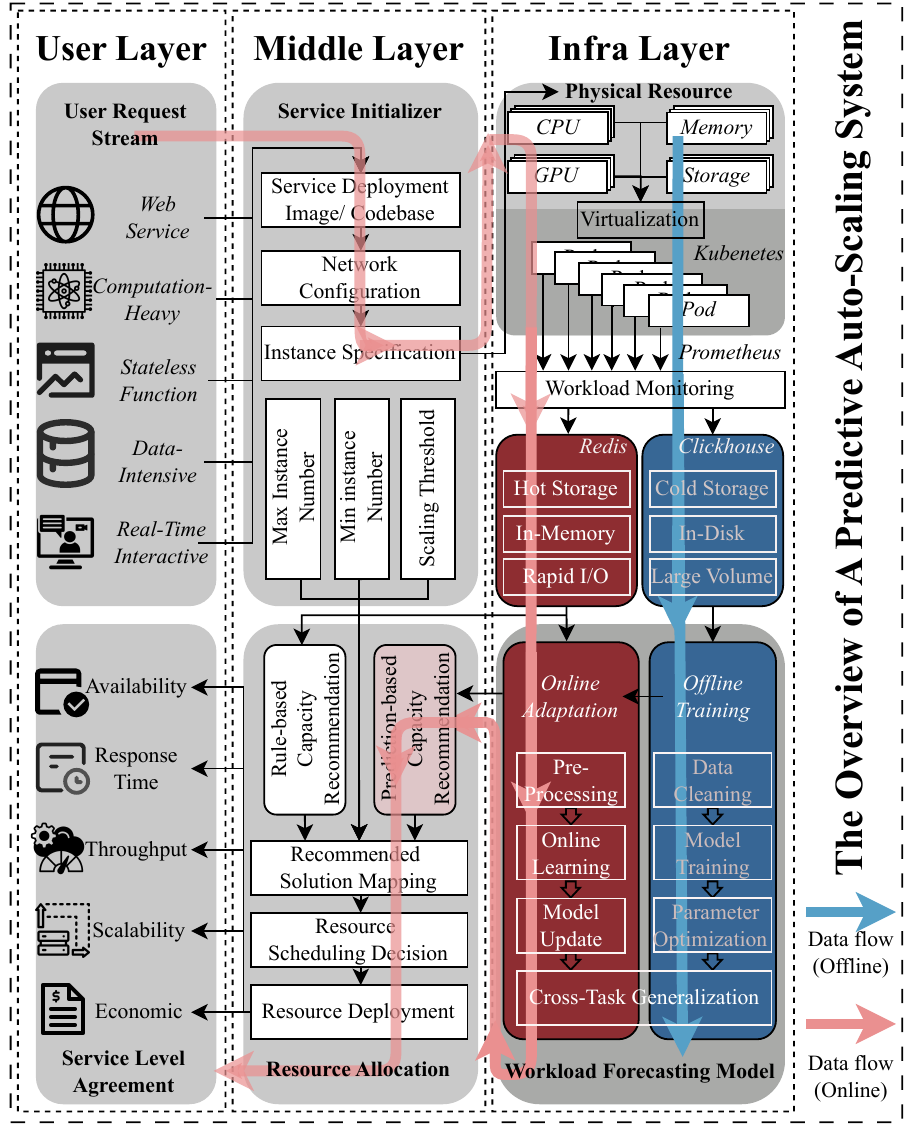}
\vspace{-1mm}
\caption{The overview of a predictive auto-scaling systems, which is divided into three layers: User Layer, Middle Layer, and Infra Layer. The system integrates predictive analytics to achieve proactive scaling decisions by a workload forecasting model. The workload forecasting model combines offline training with online adapting to help system optimize resource allocation, ensuring scalability, cost-efficiency, and adherence to service-level agreements.}% In the figure, we use solid lines to represent the forward computation process and dashed lines to represent the backward update process.}
% \vspace{-2mm}
\label{fig:intro_1}
\vspace{-3mm}
\end{figure}

Modern cloud computing systems rely on elastic auto-scaling mechanisms to dynamically allocate physical resources, balancing cost efficiency and service quality. At the core of this capability lies workload forecasting, a technique that predicts future resource demands to guide scaling decisions~\cite{poppe2022moneyball, hang2024robust, HeLTWL23}, as shown in Figure~\ref{fig:intro_1}.
Research has demonstrated that elaborate auto-scaling systems with accurate workload forecasting can enhance service quality by 20\% while reducing resource waste by 15\%~\cite{ guo2024pass}. However, to achieve the goal of cost reduction and efficiency improvement, forecasting models need to be versatile and capable of addressing a variety of challenges. Figure~\ref{fig:intro_2} shows the results that an inadequate forecast could lead to: Under-provisioning causes the system to overload, while over-provisioning incurs unnecessary resource costs~\cite{li2024flux}.
% Empirical studies indicate that suboptimal predictions may result in up to 30\% resource waste or 20\% performance degradation in typical cloud environments, underscoring the economic and operational imperative of accurate load forecasting.

\begin{figure}[]
\centering
\includegraphics[width=0.98\linewidth]{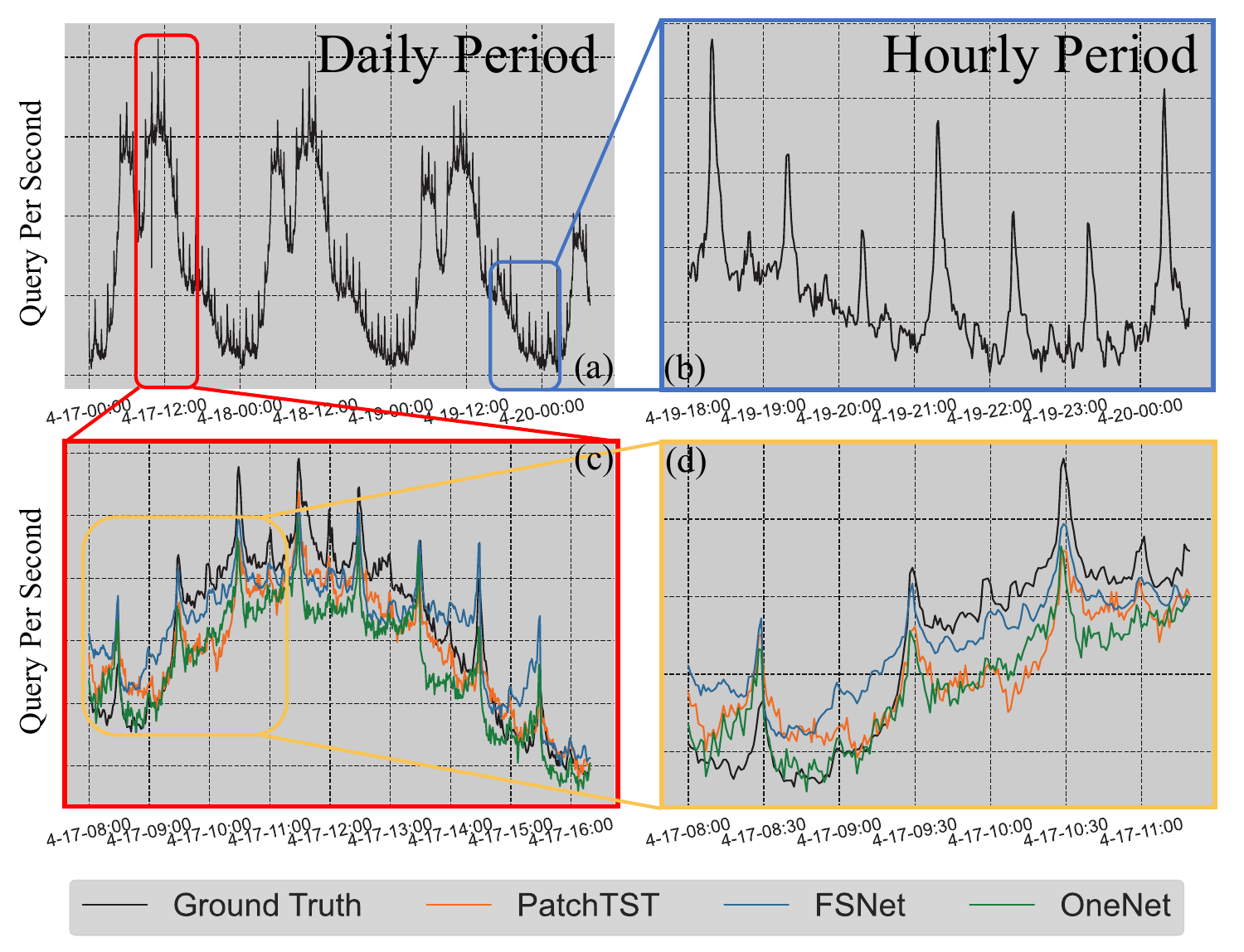}
\caption{Workload shows multi-granularity periodic pattern, and recent works cannot capture such complicated dynamics. }% In the figure, we use solid lines to represent the forward computation process and dashed lines to represent the backward update process.}
\vspace{-3mm}
\label{fig:intro_3}
\end{figure}

Cloud system workloads exhibit four distinct characteristics that complicate forecasting. \textbf{1) Complex periodic patterns} arise from multi-scale cycles (e.g., hourly, daily, weekly, and seasonal variations) intertwined with irregular bursts, as shown in Figure~\ref{fig:intro_3}. \textbf{2) Long-term prediction length} (ranging from minutes to hours) are essential due to the latency of physical resource allocation in data centers. \textbf{3) Changing dynamics} arise from either user behaviors or system updates, which necessitate that the forecasting model must be capable of online adaptation. 
% , massive-scale data streams require processing thousands of concurrent time series across distributed systems. 
% Lastly, strict real-time constraints demand millisecond-level inference latency to enable timely scaling actions (Figure 1). These challenges collectively render traditional forecasting approaches inadequate for cloud-scale deployment.
\textbf{4) Robustness} is of paramount importance, as cloud computing systems must ensure Service Level Agreements (SLAs). Therefore, as the core of the auto-scaling system, the forecasting model must be capable of handling diverse sequences and generating robust predictions.

% Current solutions fall into two categories: statistical models and deep learning approaches. Statistical methods like Exponential Smoothing (ETS) offer simplicity and interpretability but struggle with complex periodic features and long-term dependencies. While deep neural networks (e.g., TCN, Transformer) achieve higher accuracy through nonlinear pattern learning, they require extensive offline training on historical data and exhibit prohibitive computational overhead during online inference. This trade-off between adaptability and efficiency becomes particularly acute in cloud environments where workloads evolve dynamically, necessitating a paradigm shift toward online learning frameworks.
\begin{savenotes}
\begin{figure}[]
\centering
\includegraphics[width=0.98\linewidth]{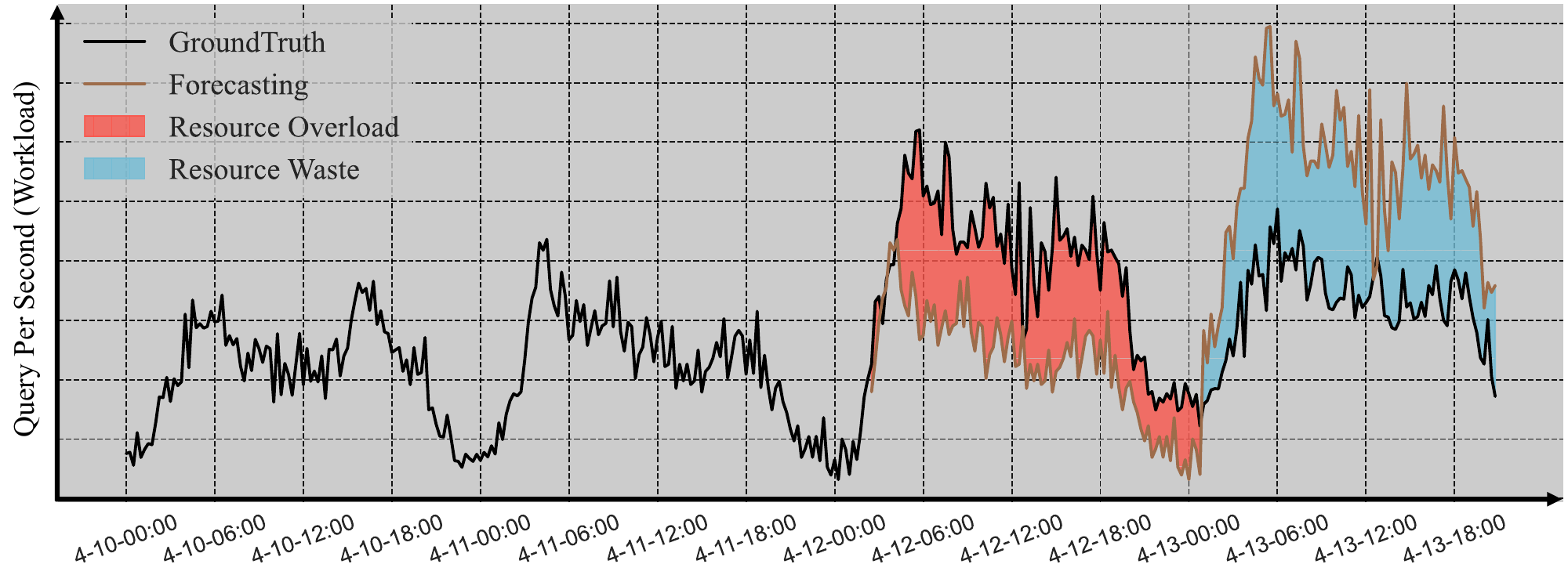}
\caption{When the predicted value exceeds the actual workload ({\color{blue}blue zone}), it not only wastes physical resources but also incurs unnecessary costs\protect\footnotemark[1]. Conversely, underestimating the workload ({\color{red}red zone}) can lead to system overload and a decrease in service quality.}% In the figure, we use solid lines to represent the forward computation process and dashed lines to represent the backward update process.}
\vspace{-2mm}
\label{fig:intro_2}
\end{figure}
\end{savenotes}
\footnotetext[1]{For a distributed computing cluster comprising hundreds of thousands of CPU cores, every 1\% of resource wastage translates to an annual revenue loss of tens of thousands, if not hundreds of thousands, of dollars.}

To address these limitations, we present an online workload forecasting model comprising four synergistic components:
\textbf{1)~Representer:} Extracts multi-resolution periodic features through multi-resolution patching operation, explicitly modeling nested cycles.
\textbf{2)~Transformer:} Leverages self-attention modules to capture long-range dependencies.
\textbf{3)~Adapter:} Implements online adaptation via cumulative gradient-based parameters updates, enabling rapid refining to workload dynamics.
\textbf{4)~Ensembler:} Combines results from different sub-networks through adaptive weighting, enhancing robustness against distribution shifts.

We also collect and open-source high-quality workload datasets\footnotemark[2] from distinct cloud service types in ByteDance Cloud, containing workload data of thousands of computing instances, spanning from 1 month to 2 months. We carefully preprocess the raw data, and make them useful tools for the community to evaluate and develop new workload forecasting methods. In online forecasting tasks on twelve datasets, compared with the previous SOTA, \model reduces forecast MSE/MAE/WMAPE by an average of \textbf{13.9\%/11.7\%/19.1\%}. In transfer forecasting tasks that simulate real-world cold-start scenarios, \model shows generalization ability and robustness, and reduces forecast MSE/MAE/WMAPE by an average of \textbf{15.3\%/14.1\%/26.3\%}.
Moreover, to substantiate the practical applicability of our model, we implement it within a real-life online kubernetes system and carry out a predictive auto-scaling test. Compared with the built-in Na\"ive HPA module in kubernetes, the scaling strategy based on \model reduces the average/maxmium Pod occupation by \textbf{7.3\%/29.4\%} and decreases the system's average/maximum latency by \textbf{5.6\%/91.7\%}. This study serves to confirm the model's efficacy and highlights its utility in real-world scenarios, underlining its potential to enhance predictive accuracy and operational efficiency in diverse applications. Our contributions are summarized as follows:

\footnotetext[2]{\url{https://huggingface.co/datasets/ByteDance/CloudTimeSeriesData}}

\begin{enumerate}
    % \item We design an online Adaptor, which serves as the foundation for creating a Transformer backbone tailored for online and transfer tasks. This novel structure enables the model to adapt more dynamically to incoming data streams, significantly improving its versatility in online scenarios.
    \item To address the challenges of online workload forecasting, we design the Rrepresenter to extract multi-granularity periodic patterns, the Adapter to achieve rapid online updates at parameter level, and the Ensembler to improve robustness with online learning theory.
    \item We develop an efficient and effective Transformer-based ensemble framework, achieving significantly more accurate results with a negligible increase in parameters.
    % \item We theoretically demonstrate that the proposed ensemble model can achieve vanishing regret.
    \item We collect and open-source workload datasets from real-world large-scale cloud systems, and conduct comprehensive experiments on them, showcasing the superiority of our proposed model over typical baselines.
    \item In Kubernetes HPA proactive auto-scaling tests, compared with the built-in Naive HPA module, \model reduces the average / maximum latency by 5.6\% / 91.7\% and the average / maximum resource consumption by 7.3\% / 29.4\%.
    % We gather data from genuine cloud computing systems and showcase the superiority of our proposed model over typical baseline approaches through comprehensive experiments. Furthermore, a detailed case study highlights the practical application and effectiveness of the model in real-world scenarios.
    
\end{enumerate}

\section{Related Works}

\textbf{{Predictive Auto-Scaling.}} Auto-scaling has emerged as a pivotal mechanism for the efficient allocation and relinquishment of resources in accordance with workload variations~\cite{baldini2017serverless}, resulting in significant reductions in operational and management expenses~\cite{qu2018auto}. 
Proactive auto-scaling based on time series forecasting can prepare resources in advance to ensure Quality of Service (QoS) during peak load periods, and reduce resources promptly during low load periods to avoid resource wastage~\cite{poppe2022moneyball, hang2024robust}. 
Zhou et al. propose AHPA~\cite{zhou2023ahpa}, a robust decomposition-based statistical method for predicting resource load in services, subsequently enabling horizontal scaling of pods based on expected resource demands. 
%This approach enhances service quality compared to the Na\"ive Horizontal Pod Autoscaler (HPA) strategy. 
Guo et al. introduce PASS~\cite{guo2024pass}, which employs a prediction framework that integrates both online and offline prediction models to enhance the accuracy of resource prediction, thereby achieving higher quality auto-scaling. 
Accurate forecasts play a vital role in such systems.
% \textbf{Overview of Forecasting Methods.}
% Early time series forecasting predominantly relied on traditional statistical methods, such as ETS \cite{holt2004forecasting}, ARIMA \cite{box1968some} and STL \cite{DBLP:journals/pvldb/HeLTWL23}. These methods are fundamental; however, they often have limitations in capturing complex temporal dynamics.
% In recent years, the emergence of deep learning has significantly propelled the field of time series forecasting, especially for long-term predictions. RNN-based methods \citep{hochreiter1997long, chung2014empirical, lai2018modeling} and CNN-based methods \citep{bai2018empirical, liu2022scinet} have shown enhanced abilities in extracting intricate patterns from time series data.
% Furthermore, Transformer-based methods \citep{zhou2021informer, kitaev2020reformer, zhang2022crossformer, nie2022time} can uncover the temporal dependencies between time points through the attention mechanism.
% Additionally, MLP-based methods \citep{oreshkin2019n, challu2022nhits, zeng2023transformers} offer a straightforward yet effective approach for learning the non-linear temporal dependencies in time series.

\noindent\textbf{Workload Forecasting in Systems.}
In large-scale systems, workload modeling is usually done with statistical or machine learning methods~\cite{calheiros2014workload,das2016automated,chen2020adaption,higginson2020database}. With the development of deep learning, state-of-the-art time series prediction techniques for workloads have begun to be applied to system load balancing, parameter optimization, capacity scaling, etc. LoadDynamics~\cite{jayakumar2020self} employs a combination of LSTM and Bayesian optimization for handling the dynamic fluctuations of cloud workloads. Seagull~\cite{Seagull} is an end-to-end generic ML infrastructure for load prediction and optimized resource allocation by adjusting plugged ML models. WGAN-gp Transformer~\cite{ArbatJL0K22} adopts a Transformer network as a generator and a multi-layer perception as a critic to predict cloud workload. QueryBot~\cite{ma2018query} combines linear regression and RNNs for various database workload patterns. DBAugur~\cite{DBAugur} uses adversarial neural networks to predict the trends of database workloads and shows the superiority of index selection and data region migration tasks. 

\begin{table}[!htp]
    \caption{Workload Metrics by Architecture Layers.}
    \label{tab:workload}
    \centering
    \setlength{\tabcolsep}{3pt}
    \small % 9pt字号匹配VLDB
     \begin{tabular}{@{}l@{\,}ccc@{\,}c@{}}
        \toprule
        \rowcolor{layergray}
        \multicolumn{1}{c}{\textbf{Service Layer}} & 
        \multicolumn{3}{c}{\textbf{Network}} & 
        \textbf{App Service} \\
        \cmidrule(r){2-4} \cmidrule(l){5-5}
        \textit{Throughput \&} & In-Traffic & Out-Traffic & TCP Conn. & Requests \\
        \textit{\quad Reliability} & Latency & UDP Loss & Packet Rate & Resp. Time \\
        \midrule[0.3pt] % 层间分隔
        \rowcolor{layergray}
        \multicolumn{1}{c}{\textbf{System Layer}} & 
        \multicolumn{2}{c}{\textbf{Database}} & 
        \multicolumn{2}{c}{\textbf{OS}} \\
        \cmidrule(r){2-3} \cmidrule(l){4-5}
        \textit{Performance \&} & Conn. Count & Cache Hit & Waiting Procs & Running Procs \\
        \textit{\quad Latency} & Query Time &  Index Usage & Context Switches & Load \\
        \midrule[0.3pt]
        \rowcolor{layergray}
        \multicolumn{1}{c}{\textbf{Infra Layer}} & 
        \multicolumn{2}{c}{\textbf{CPU}} & 
        \textbf{Memory} & 
        \textbf{Storage} \\
        \cmidrule(r){2-3} \cmidrule(lr){4-5}
        \textit{Core Metrics \&} & Total Usage & Core Load & Usage & Disk I/O \\
        \textit{\quad Health} & Temperature & Frequency & Swap Rate & Remaining \\
        \bottomrule
    \end{tabular}
    \vspace{-2mm}
    % \caption*{\footnotesize Note: 3层架构划分，40\%空间节省，适配VLDB单栏（398pt）}
\end{table}

\section{Preliminaries}

\subsection{Workload Date Stream}

In the context of cloud computing, a workload series represents a time-ordered aggregation of workloads generated by jobs or applications operating on cloud infrastructure~\citep{alahmad2021proactive}. Table \ref{tab:workload} concludes common types of workload in cloud environment, including distinct resource-consumption patterns (such as CPU and memory usage) and user-request metrics (like response time). In online scenarios, the workloads appear as potentially unbounded streams of continuous data in real (or near-real) time, reflecting system statuses, as described in Def.~\ref{def:stream}. 

\begin{myDef}[Workload Stream] \label{def:stream}
A workload stream, or an kind of online time series, is a sequence of signals (such as CPU usage, memory usage, etc.), measured at successive time stamps, which are assumed to be spaced at uniform intervals (such as every minute). We denote by $x_t$ the signal recorded at time stamp $t$, and a data stream can be denoted as $\mathbf{X}=\{x_1, x_2,\cdots\,x_t,\cdots\}$. 
\end{myDef}

With the operation of the large-scale system, new workload metrics are continuously recorded, and the stream is continuously longer. Due to the need for real-time and efficiency, we cannot use unlimited memory and storage to record all histories and then predict the future.
As shown in Figure~\ref{fig:intro_1}, online workload streams are stored in two forms: cold storage and hot storage. Cold storage uses hard disks as the storage medium and is managed through OLAP databases such as ClickHouse. It can store large-scale workload data on a monthly basis, but its read speed is limited, making it unsuitable for real-time analysis tasks. Hot storage, on the other hand, uses memory as the storage medium and is managed through in-memory databases such as Redis. Its read and write speeds are highly efficient, supporting real-time analysis. However, due to capacity constraints, it can only store a limited amount of data, such as workload from the past week.

Given the limited historical data accessible in online environments, which is insufficient for complete offline training, a pipeline combining offline pretraining with online updating is more suitable. The model initially learns from large-scale historical data and then adapts to the online environment based on the limited available data to generate accurate and efficient predictions.
% Therefore we are considering an \emph{online workload  forecasting} problem over workload stream $\mathbf{X}$.

% For a computing instance, the workload series can be mathematically defined as $\mathbf{X}=\{x_0,x_1,\dots,x_{L-1}\}$, where $L$ denotes the length of a given time period, which is equivalent to the number of time steps utilized as input, and $w_i$ represents the numerical value of the specific workload at time-step $i$.

\subsection{Online Workload Forecasting}

Firstly, there is no separation of training and evaluation in an online setting. Instead, learning occurs over a sequence of time-steps~\cite{phamlearning}. Hence, At each time stamp $t$, we can define a length parameter $L$ and use observation history $\textbf{X}_t=\left\{\mathbf{x}_{t-L},\mathbf{x}_{t-L+1},\cdots,\mathbf{x}_{t-1}\right\}=\{\mathbf{x}\}_{t-L}^{t-1}$ , where $\mathbf{x}_t\in \mathbb{R}^M$ denotes the $M$ kinds of workload metrics recorded at time $t$, to produce a $H$-length prediction $\mathbf{\hat{Y}}_t=f(\textbf{X}_t;\theta^f_t)=\{\hat{\mathbf{x}}_{L},\hat{\mathbf{x}}_{L+1},\dots,\hat{\mathbf{x}}_{L+T-1}\}$ with prediction model $f$ and associated parameter set $\theta^f_t$. $\textbf{X}_t$ is a fixed-length observed series history, focusing on the latest and limited observation in online scenarios.

We use loss function $\ell(\hat{\mathbf{Y}}_t,\mathbf{Y}_t)$ to measure the gap between the prediction $\hat{\mathbf{Y}}_t$ and truth $\mathbf{Y}_t$. Specifically, $\ell$ maps the prediction and the truth into a real number in $[0,1]$, and the more accurate the prediction is, the smaller the loss is. Depending on the requirements, there are many choices of $\ell$. 

Secondly, since a data stream is recorded continuously, we notice that the learning parameter $\theta$ should be updated at each time $t$, which derives the formal definition of OnPred, denoted as Def.~\ref{def:onpred}.

\begin{myDef}[Online Workload Forecasting]\label{def:onpred}
At time stamp $t$, given an $L$-length history stream $\mathbf{X}_t$, the online workload forecasting task requires a prediction $\hat{\mathbf{Y}}^f_t=f(\mathbf{X}_t;\theta^f_t)$. After making the prediction, the prediction model $f$ then receives the true answer $\mathbf{Y}_t$ and generates a related loss $\ell \left(\mathbf{\hat{Y}}_t, \mathbf{Y}_t\right)$. Whenever the loss is nonzero, $f$ updates the parameter set from $\theta^f_t$ to $\theta^f_{t+1}$ by applying some update strategy on the training example pair $(\mathbf{X}_t, \mathbf{Y}_t)$. By running the online learning $T$ time-steps, the goal of online workload forecasting is to minimize the cumulative loss, i.e., $\sum_{t=1}^T \ell (\hat{\mathbf{Y}}^f_t, \mathbf{Y}_t)$.
\end{myDef}

PROCEED~\cite{zhao2024proactive} proposes the view that an online learning task inherently has a \( H \)-step feedback delay in time series forecasting, resulting in a temporal gap (at least \( H \) steps) between available training samples and the test sample. In practical workload forecasting tasks, there is typically a certain interval between two consecutive predictions, as repeated predictions would lead to resource waste. Ideally, this interval should be set to the length of the forecast horizon to meet the needs of guiding scaling operations. Therefore, the online forecasting setting is immune to this temporal gap.

% Given the historical value of a workload series $\mathbf{X}=\{x_0,x_1,\dots,x_{L-1}\}$,  the objective of workload forecasting is to predict the future workload series $\mathbf{\hat{X}}=\{x_{L},x_{L+1},\dots,x_{L+T-1}\}$. In this expression, $T$ represents the length of the forecasting horizon, that is, the number of time steps to be predicted.
% % In this paper, we treat workload series as multivariate series for the purpose of forecasting.
% The target of workload forecasting is to make the prediction $\hat{X}$ as accurate as possible, which is equivalent to minimizing the gap between the prediction $\hat{X}$ and the corresponding ground truth value $Y$. The accuracy of forecasting can be measured by several metrics such as MSE, MAE and SMAPE. The smaller these metrics are, the more accurate the forecasting result is.

\begin{table}
\caption{Symbols and Notations}\label{tab:symbol}
\begin{tabular}{cc}
\toprule Symbols & Definitions \\
\midrule 
$\mathbf{X}$ & workload data stream  \\ %$\mathbf{X}=\{\mathbf{x}_1,\mathbf{x}_2,\cdots\,\mathbf{x}_i,\cdots\}$\\ 
$\mathbf{x}_t$ &  workload series point at timestamp $t$\\
$M$ & number of metrics of the workload stream\\
$L$ & lookback length \\
$H$ & forecasting length\\
$\mathbf{X}_t, \mathbf{Y}_t$ & input and output series at time $t$\\
$\hat{\mathbf{Y}}_t=f(\mathbf{\mathbf{X}}_t;\theta_t^f)$ & forecasting with input $\mathbf{X}_t$ and parameter $\theta_t^f$ \\
$P$ & patch size \\
$D, D^\prime, D_{ff}$ & hidden dimension of different modules\\
$W^O, W^Q, W^K, W^V$ & learnable parameter matrices \\
$\nabla$ & cumulative gradient \\
$\mathbf{\alpha, \beta}$ & adaptation coefficients \\
$\triangle, \Pi$ & decision space for model ensemble  \\
$\mathbf{w}, \mathbf{\pi}$ & decision strategy for model ensemble \\

\bottomrule
\end{tabular}
\end{table}

In general, the performance of an online model is measured by comparing its cumulative loss with the minimum cumulative loss of the offline algorithm, and its difference is denoted as \emph{regret}~\cite{hoi2021online}:
\begin{equation}\label{eqn:regret}
regret = \sum_{t=1}^T \ell (f(X_t;\theta^f_t), x_t) - \inf_{\theta}\sum_{t=1}^T \ell (f(X_t;\theta), x_t).
\end{equation}

The second term in Eqn.~\eqref{eqn:regret} is the loss suffered by the model with the best parameter setting $\theta^*$, which can only be known in hindsight after seeing the full data stream. An online model's regret is \emph{sub-linear} as a function of $T$, if and only if $regret = o(T)$, which implies $\lim\limits_{T\rightarrow \infty} \dfrac{regret}{T} \rightarrow 0$ and thus on average the model performs almost as well as setting the best parameter in hindsight. We always expect an online learner achieving a sub-linear regret. Table~\ref{tab:symbol} presents the primary notations with the corresponding definitions.

\begin{figure}[!htp]
\centering
\includegraphics[width=0.98\linewidth]{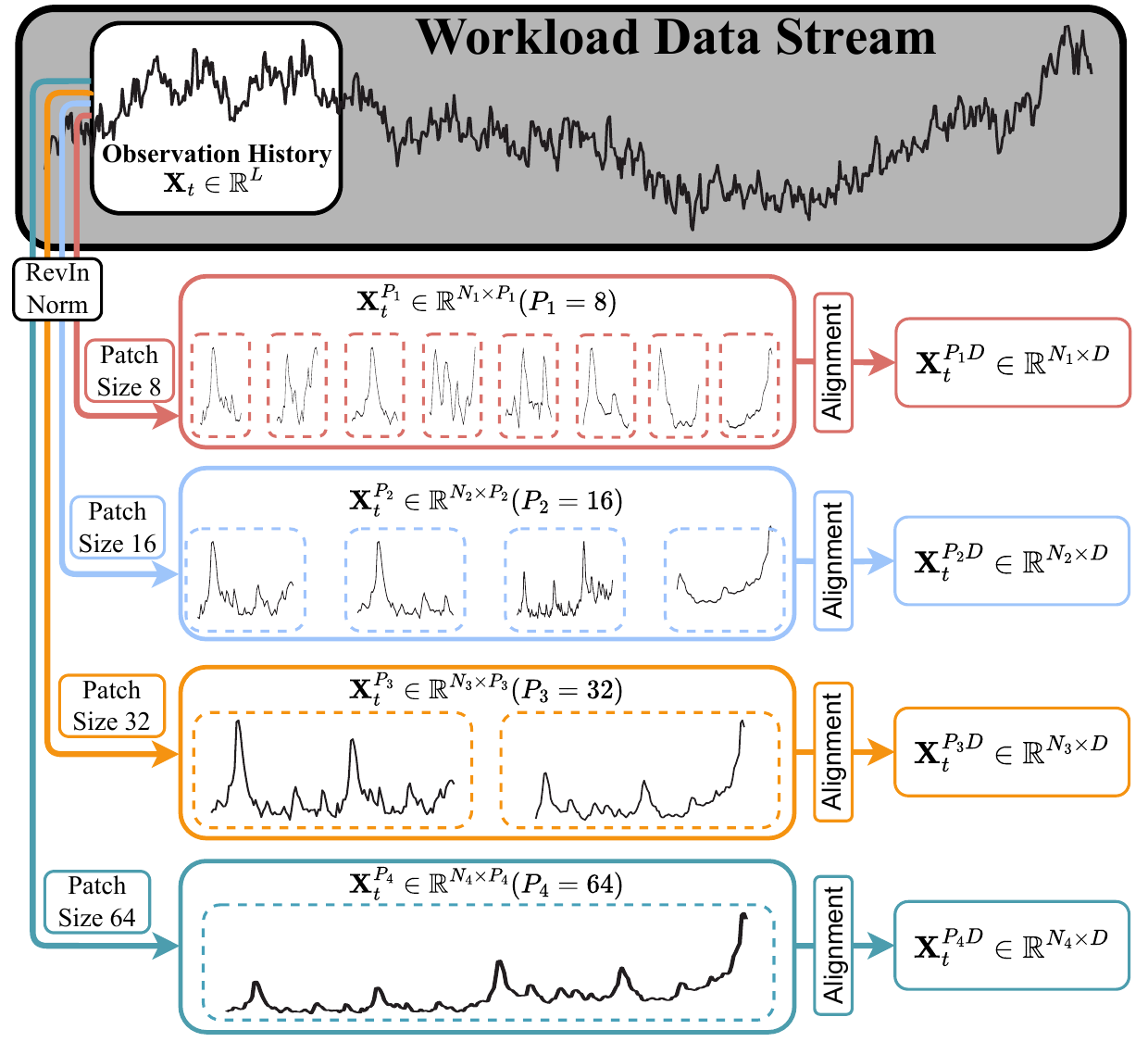}
\caption{The pipline of Representer. Initially, the RevIN Norm technique is employed to regularize the data distribution of the input historical sequence. Subsequently, the sequence undergoes a patching operation using patches with different lengths (8, 16, 32, and 64), which segments the sequence into equal-length patches. Finally, a linear layer is utilized to map the last dimension of these patches to a uniform hidden layer dimension D, ensuring consistency for further processing in the model.}% In the figure, we use solid lines to represent the forward computation process and dashed lines to represent the backward update process.}
% \vspace{-2mm}
\label{fig:representer}
\vspace{-4mm}
\end{figure}

\section{Model Design}
In this section, we outline the methodologies designed to address the challenges in workload forecasting. To extract complex periodic patterns, we designed the Reresenter, which centers on a multi-resolution patching operation. To model long-term dependencies, we leverage a Transformer forecasting model. To capture changing dynamics, we design the Adapter module to achieve effective online parameter updates. To generate robust predictions, we design an Ensembler with two alternative ensemble modules (Online Scaling and Follow-The-Perturbed-Leader) based on online learning theory. Through the multi-resolution patching operation, the input is transformed into multiple sets of patches, retaining periodic information at various levels. These sets of patches are then fed into the parameter-shared Transformer to produce corresponding multiple outputs. During the offline training phase, each output is used independently to train the corresponding sub-networks. In the online inference phase, the Ensembler integrates these multiple outputs to generate the final prediction results. Meanwhile, the Adapter ensures the model fits the changing dynamics of the workload.

\subsection{Representer}

We designed the Representer module to extract multi-level periodic features from workload series and represent them in a form suitable for parallel processing. Specifically, the Representer has four operations: channel independence, RevIN, multi-resolution patching, and alignment.

\subsubsection{Channel Independence}

In channel independence setting~\cite{nietime}, each channel of a multivariate time series is treated as independent. Specifically, the multivariate time series $\mathbf{X}_t\in \mathbb{R}^{L\times M}$ is organized as a $M$-sized mini-batch of single-variate time series $\{{\mathbf{X}_{t,i}\}}_{i=1}^M$. Through channel independence, we can fully utilize the parallelism of GPUs, expand the training set by $M$ times, and avoid overfitting caused by complex inter-channel relationships. Since each series in the minibatch goes through the same parallel pipeline, we abbreviate $\mathbf{X}_{t,i}$ as $\mathbf{X}_t\in\mathbb{R}^L$.%This allows the model to focus on the unique patterns of each variable, avoiding overfitting caused by complex inter-channel relationships. 

\subsubsection{RevIN}

Reversible Instance Normalization (RevIN)~\cite{kim2021reversible} is a normalization method for time series. It performs instance normalization by calculating the mean \(\mu\) and standard deviation \(\sigma\) over $\mathbf{X}_t$. Then, it normalizes the data:
% \[
% \mu = \frac{1}{T} \sum_{i=t+1}^{t+L} x_i, \quad \sigma = \sqrt{\frac{1}{T} \sum_{t=1}^{T} (\mathbf{X}_t - \mu)^2 + \epsilon}
% \]
$
\hat{\mathbf{X}}_t = \frac{\mathbf{X}_t - \mu}{\sigma}.
$
Later, RevIN applies an affine transformation with learnable parameters \(r_1\) and \(r_2\):
\(\hat{\mathbf{X}}_t = r_1 \hat{\mathbf{X}}_t + r_2.
\)
During denormalization, the process is reversed to restore the original data distribution.

\subsubsection{Multi-resolution patching}
The patching operation can help extract the local periodic information from a series. Specifically, given the patch size $P$, the input time series $\hat{\mathbf{X}}_t$ can be divided into multiple $P$-sized patches $\mathbf{X}^P_t\in\mathbb{R}^{N\times P}$, where $N$ is the number of patches calculated as $N=\lceil L / P\rceil$.
% For example, a time series is partitioned into patches $\mathbf{X}^P_t\in\mathbb{R}^{N\times P}$, 
To handle the end of the series, the last value of $\mathbf{X}_t$ is padded with repeated numbers $(L\mod P)$, ensuring that each patch maintains consistent dimensions. 
%Different patch sizes are employed to capture diverse characteristics of the time series, with larger patches capturing high-frequency details and smaller patches retaining a larger receptive field for low-frequency features, thereby optimizing computational efficiency while preserving essential patterns.
Research shows that the patch-based projections with
larger patch sizes can capture high-frequency features and vice versa~\cite{woounified}.
Furthermore, in order to represent the periodic features of different levels, we design a multilevel patching method. We utilize a group of different patch sizes $\{P_1, \cdots, P_d\}$ and process the input time series into multiple groups of patches $\left\{\mathbf{X}^{P_i}_t\right\}_{i=1}^d$.

%We use larger patch sizes for capturing high-frequency characteristics, effectively reducing the computational burden associated with the quadratic costs of attention mechanisms, and use smaller patch sizes to keep larger receptive field for low-frequency characteristics~\cite{woounified}.

\subsubsection{Alignment}

Using $d$ independent models to learn and predict across $d$ groups of inputs incurs an additional computational cost that scales linearly with $d$. To address this efficiency challenge, we employ the multi-input multi-output (MIMO) mechanism\cite{havasitraining}, utilizing a single forecasting model to independently learn from each of the $d$ input groups and produce $f$ outpus. This approach requires input alignment, which we achieve by projecting each input group through a separate linear layer to a shared dimension:
% Using $d$ independent models to learn and predict in $d$ groups of input incurs $d$ times the additional cost. Therefore, we drew on the idea of the multi-input multi-output (MIMO) mechanism, employing the same forecasting model to independently learn from each of the d groups of inputs. This approach necessitates the alignment of the inputs. Each input group is projected through a separate linear layer to align them to the same dimension, which aligns their dimensions in the last axis 
$\mathbf{X}^{P_iD}_t = Linear(\mathbf{X}_t^{P_i}) \in \mathbb{R}^{N_i \times D}$, where $D$ is the hidden dimension. As each group of patches $\mathbf{X}_t^{P_i}$ will be input in parallel into the subsequent pipeline, and we abbreviate $\mathbf{X}_t^{P_i}$ as $\mathbf{X}^P_t$.

% Meanwhile, to align the inputs so they can be fed into the same forecasting backbone, we separately conduct RevIn normalization~\cite{kim2021reversible} and linear embedding on each group of patches. 
% This aligns their dimensions in the last axis $\mathbf{X}_{P_i} \in \mathbb{R}^{M\times N_i \times D}$, where $D$ is the hidden dimension. 

\subsection{Transformer}

We introduce a Transformer module to empower the model with long-term forecasting ability. 
With the Embedding matrices \\ $\mathbf{W}^Q, \mathbf{W}^K, \mathbf{W}^V \in \mathbb{R}^{D\times D^\prime}$, the input $\mathbf{X}^{P}_t$ is embedded into $\mathbf{Q}, \mathbf{K}, \mathbf{V}\in \mathbb{R}^{P\times D^\prime}$, where $\mathbf{Q} = \mathbf{X}^{P}_t\mathbf{W}^Q$, $\mathbf{K} = \mathbf{X}^{P}_t\mathbf{W}^K$, $\mathbf{V} = \mathbf{X}^{P}_t\mathbf{W}^V$, and $D^\prime$ is denoted as the hidden dimension for multi-head attention computation. 
After the Representer module and with positional encoding applied, the input is fed into an attention block with several attention layers.
The multi-head self-attention (MHSA) mechanism allows the model to jointly attend to information from different representation subspaces at various positions:
\begin{equation}  
\begin{aligned}  
\label{eq:self-attention}
MHSA(\mathbf{X}^{P}_t) = \ Concat&(head_1,\dots, head_h)\mathbf{W}^O, \\
\text{where} \ head_i = & Softmax(\frac{\mathbf{Q}_i\mathbf{K}_i^T}{\sqrt{D^\prime}})\mathbf{V_i}, \\
\mathbf{Q}_i = \mathbf{X}^{P}_t\mathbf{W}^Q_i,\  \mathbf{K}_i &= \mathbf{X}^{P}_t\mathbf{W}^K_i,\  \mathbf{V}_i = \mathbf{X}^{P}_t\mathbf{W}^V_i,
\end{aligned}  
\end{equation}  
where the projections are parameter matrices $\mathbf{W}_i^Q \in \mathbb{R}^{D\times D^\prime}, \mathbf{W}_i^K \in \mathbb{R}^{D\times D^\prime}, \mathbf{W}_i^V \in \mathbb{R}^{D\times D^\prime}, \mathbf{W}^O\in\mathbb{R}^{hD^\prime\times D}$ and $D^\prime$ denotes the hidden dimension for multi-head attention computation. 

\begin{figure}[!htp]
\centering
\includegraphics[width=0.98\linewidth]{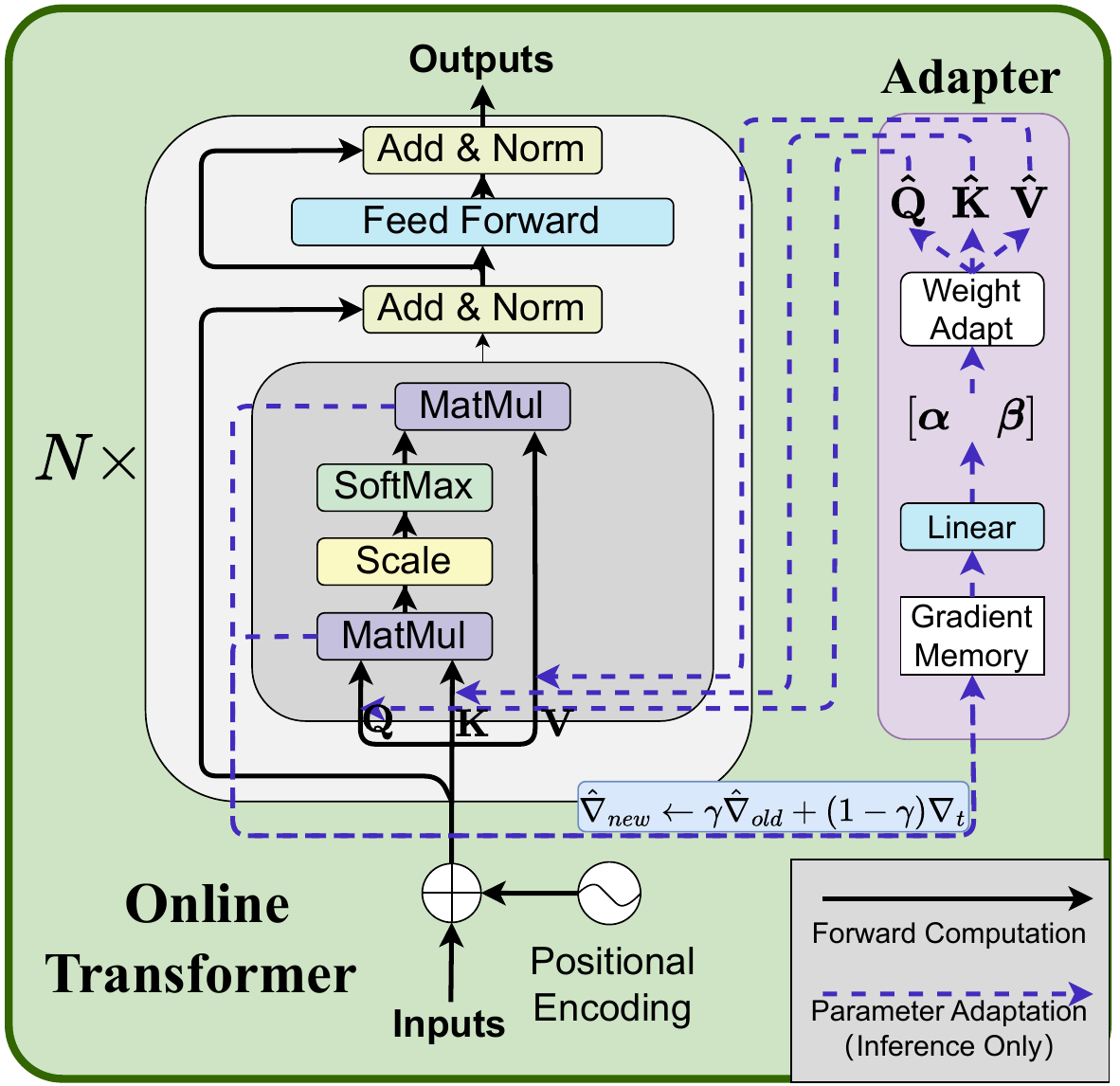}
\caption{The structure of Online Transformer with Adapter. During the training phase, the Online Transformer's computation process aligns with that of the original Transformer Encoder, including self-attention layers, skip connections, batch normalization, and a feed-forward network (as shown by the solid \textbf{black} line). In the online inference phase, the Adapter adjusts the weights of the self-attention layer at the parameter level, enabling the model to dynamically adapt to data changes (as shown by the dashed {\color{blue}blue} line).}% In the figure, we use solid lines to represent the forward computation process and dashed lines to represent the backward update process.}
% \vspace{-2mm}
\label{fig:transformer}
\vspace{-3mm}
\end{figure}

We construct an attention layer with MHSA, skip-connection, Batch Normalization~\cite{ioffe2015batch}, and a feed-forward network (FFN): 
\begin{gather}  
\mathbf{X}_A = BatchNorm(\mathbf{X}^{P}_t + MHSA(\mathbf{X}^{P}_t)),\\
FFN(\mathbf{X}_A) = \max(0, \mathbf{X}_A\mathbf{W}_1 +b_1)\mathbf{W}_2 + b_2,\\
\mathbf{X}_{O} = BatchNorm(\mathbf{X}_A + FFN(\mathbf{X}_A)),
\end{gather}
where $\mathbf{W}_1\in\mathbb{R}^{D \times D_{ff}}$, $\mathbf{W}_2\in\mathbb{R}^{D_{ff} \times D}$, $b_1\in\mathbb{R}^{D_{ff}}$, and $b_2\in\mathbb{R}^{D}$. The output of an attention layer is fed into the next attention layer as input. Finally, 
%the last attention block produces $\mathbf{X}_{output}\in\mathbb{R}^{P\times D}$. 
We flatten output from the last attention block $\mathbf{X}_{O} \in\mathbb{R}^{P\times D}$ and map it to the final forecasting $\mathbf{Y}_t = f(\mathbf{X})\in\mathbb{R}^{H}$ through a linear layer $\mathbf{Y}_t = {Linear}(\mathbf{X}_{O})$.
%(due to the setting of channel independence, the model produces outputs for all channels $f(\mathbf{X})\in\mathbb{R}^{M\times H}$ simultaneously).

\subsection{Adapter}

In the context of online training, the inherent noise and non-stationarity of time series data can lead to significant fluctuations in the gradient of a single sample, thereby introducing noise into the adaptation coefficients. 
To mitigate this issue, we draw inspiration from the ideas of FSNet~\cite{phamlearning} and employ the Exponential Moving Average (EMA) of the attention layers' gradient. 
This approach effectively smoothes the noise in online training, enabling efficient capture of the temporal information inherent in time series data:
% $$
% \hat{\boldsymbol{g}}_l \leftarrow \gamma \hat{\boldsymbol{g}}_l+(1-\gamma) \boldsymbol{g}_l^t,
% $$
\begin{equation}
    \hat{\nabla}_{new} \leftarrow \gamma \hat{\nabla}_{old}+(1-\gamma) \nabla_t,
\end{equation}
where $\nabla_t$ denotes the gradient of the attention layer at time $t$ and $\hat{\nabla}$ denotes the EMA gradient. The adaptor takes $\hat{\nabla}$ as input and maps it to the adaptation coefficients $\boldsymbol{\alpha, \beta}$. We adopt the element-wise transformation as the adaptation process thanks to its simplicity and promising results in continual learning~\cite{yin2021mitigating, phamlearning}. 
Particularly, the adaptation coefficients consist of two components: a weight adaptation coefficient $\boldsymbol{\alpha}$ and a feature adaptation coefficient $\boldsymbol{\beta}$.%, concatenated together as $\boldsymbol{u}_l=\left[\boldsymbol{\alpha}_l ; \boldsymbol{\beta}_l\right]$. % We also absorb the bias transformation into $\alpha_l$ for brevity.
The adaptation process for an attention layer involves two steps: weight and feature transformations. First, we compute two adaptation coefficients by a linear layer fed with the flattened EMA gradient,
\begin{equation}
{\left[\boldsymbol{\alpha}, \boldsymbol{\beta}\right] } =Linear\left(Flatten(\hat\nabla)\right).
\end{equation}
Then, we apply weight adaptation to the parameter matrix $\mathbf{W}^Q, \mathbf{W}^K, \mathbf{W}^V$ and feature adaption the hidden feature embeddings, respectively. Taking query embedding as an example:
% $$
% \begin{aligned}
% {\left[\boldsymbol{\alpha}_l, \boldsymbol{\beta}_l\right] } &=\boldsymbol{u}_l, \text { where } \boldsymbol{u}_l=\boldsymbol{\Omega}\left(\hat{\boldsymbol{g}}_l ; \boldsymbol{\phi}_l\right) \\
% \tilde{\boldsymbol{\theta}}_l =\operatorname{tile}\left(\boldsymbol{\alpha}_l\right) \odot \boldsymbol{\theta}_l, & \text { and } \tilde{\boldsymbol{h}}_l=\operatorname{tile}\left(\boldsymbol{\beta}_l\right) \odot \boldsymbol{h}_l, \text { where } \boldsymbol{h}_l=\tilde{\boldsymbol{\theta}}_l \circledast \tilde{\boldsymbol{h}}_{l-1} .
% \end{aligned}
% $$

% \begin{gather}
% \tilde{{\mathbf{W}}}^K =\boldsymbol{\alpha}_l\odot {{\mathbf{W}}}^K, \label{eq:adaptor}\\
% \mathbf{Q}=\mathbf{X}^{P}_t\tilde{\mathbf{W}}^K_i,\\
%  \tilde{\mathbf{Q}}=\boldsymbol{\beta}_l \odot \mathbf{Q}.
% \end{gather}
\begin{equation}  
\begin{aligned} 
\label{eq:adaptor_full}
\tilde{{\mathbf{W}}}^Q &= \boldsymbol{\alpha}_l \odot {{\mathbf{W}}}^Q, \\  
\mathbf{Q} &= \mathbf{X}^{P}_t \tilde{\mathbf{W}}^Q_i, \\  
\tilde{\mathbf{Q}} &= \boldsymbol{\beta}_l \odot \mathbf{Q}.  
\end{aligned}  
\end{equation}

Here, $\mathbf{W}^Q$ is the query embedding weight matrix, $\tilde{{\mathbf{W}}}^Q$ denotes the adapted weight, $\odot$ denotes the element-wise multiplication. Adapted query embedding $\hat{\mathbf{Q}}$ will be used in Eq.(\ref{eq:self-attention}) replacing $\mathbf{Q}$.

\subsection{Ensemler}

To generate robust predictions, it is necessary to integrate the forecasting results produced based on different levels of feature inputs. Accordingly, we have designed two ensemble module, Online Scaling and Follow The Perturbed Leader, of which the former achieves better performance while the other achieves better efficiency.

\subsubsection{Online scaling: Choose the best expert with online convex optimization}

%For notation clarity, here we denote $\mathbf{x} \in \mathbb{R}^{L \times M}$ as the historical data, $\mathbf{y} \in \mathbb{R}^{H \times M}$ as the forecast target. Our current method involves the integration of multiple complementary models. Therefore, how to better integrate model predictions in the online learning setting is an important issue. 
Exponentiated Gradient Descent (EGD)~\cite{kivinen1997exponentiated} is a commonly used ensemble method. Specifically, the decision space $\triangle$ is a $d$-dimensional simplex, i.e. $\triangle=\left\{\mathbf{w}_t \mid w_{t, i} \geq 0\right.$ and $\left.\left\|\mathbf{w}_t\right\|_1=1\right\}$, where $t$ is the time step indicator and we omit the subscript $t$ for simplicity when it's not confusing. Given the online data stream $\mathbf{X}$, its forecasting target $\mathbf{Y}$, and $d$ forecasting experts with different parameters (in \model there are $d$ independent subnetworks with a part of shared parameters) $\mathbf{F}=\left[f_i(\mathbf{X})\right]_{i=1}^d$, the player's goal is to minimize the forecasting error as
\begin{equation}
\min _{\mathbf{w}} \left\|\sum_{i=1}^d w_i f_i(\mathbf{X})-\mathbf{Y}\right\|^2 \quad \text { s.t. } \quad \mathbf{w} \in \triangle
\end{equation}

According to EGD, choosing $\mathbf{w}_1=\left[w_{1, i}=1 / d\right]_{i=1}^d$ as the center point of the simplex, the updating rule for each $w_i$ will be
% $$
% w_{t+1, i}=\frac{w_{t, i} \exp \left(-\eta\left\|f_i(\mathbf{X})-\mathbf{Y}\right\|^2\right)}{Z_t}=\frac{w_{t, i} \exp \left(-\eta \ell_{t, i}\right)}{Z_t}
% $$
\begin{equation}
    w_{t+1, i}=\frac{w_{t, i} \exp \left(-\eta\left\|f_i(\mathbf{X})-\mathbf{Y}\right\|^2\right)}{Z_t}
\end{equation}
where $Z_t=\sum_{i=1}^d w_{t, i} \exp \left(-\eta \left\|f_i(\mathbf{X})-\mathbf{Y}\right\|^2\right)$ is the normalizer, and the algorithm can achieve the regret bound that is sublinear in $T$, the total time steps, signaling proficient long-term performance.

\begin{myTheo}\label{thm:egd}
(Online convex optimization bound~\cite{ghai2020exponentiated}) For $T>2 \log (d)$, the EGD policy admits a vanishing regret
% \begin{equation}
% \sum_{t=1}^T \mathcal{L}\left(\mathbf{w}_t\right)-\inf _{\mathbf{u}} \sum_{t=1}^T \mathcal{L}(\mathbf{u}) \leq \sum_{t=1}^T \sum_{i=1}^d w_{t, i}\left\|f_i(\mathbf{x})-\mathbf{y}\right\|^2-\inf _{\mathbf{u}} \sum_{t=1}^T \mathcal{L}(\mathbf{u}) \leq \sqrt{2 T \log (d)}
% \end{equation}
\begin{equation}
Regret = \sum_{t=1}^T \mathcal{L}\left(\mathbf{w}_t\right)-\inf _{\mathbf{u}} \sum_{t=1}^T \mathcal{L}(\mathbf{u}) = O(\sqrt{T})
%&\leq \sum_{t=1}^T \sum_{i=1}^d w_{t, i}\left\|f_i(\mathbf{x})-\mathbf{y}\right\|^2-\inf _{\mathbf{u}} \sum_{t=1}^T \mathcal{L}(\mathbf{u}) 
% \leq \sqrt{2 T \log (d)}
\end{equation}
\end{myTheo}

Nonetheless, it's a well-acknowledged issue that an exponentially weighted average forecaster tends to react sluggishly to significant shifts in data distribution. This response lag is often discussed in the context of online learning circles as the "slow switch phenomenon"~\cite{cesa2006prediction}. Inspired by related algorithms~\cite{blum2007external, wen2024onenet} developed to address this issue in online learning, we adopt the idea of finding an activation function that maps the original policy $\mathbf{w}_t$ to a new one based on the recent loss of all experts.

Given the latest observed ground truth $\mathbf{Y}$, the set of corresponding $d$ forecastings $\mathbf{F}=\{f_i(\mathbf{X})\}_{i=1}^d$ and the EGD weights $\mathbf{w} = \{w_i\}_{i=1}^d$, we employ an Online Scaling module to learn from the feedback and dynamically adjust the next forecasting, ensuring a better fit with the actual series. We name this method \modelEGD. The Online Scaling module consists of an input embedding layer, a multi-head self-attention layer equipped with an Online Adaptor and an output linear layer. Input embedding layer stacks the forecastings and the ground truth together and maps the time dimension $H$ to the hidden dimension $D$ with a linear layer. After the input embedding layer, $\mathbf{F}\in\mathbb{R}^{d\times D}$  undergoes self-attention calculation, the process of which is described in Eq.(\ref{eq:self-attention}) and Eq.(\ref{eq:adaptor_full}).
% \begin{gather*}  
% MHSA(\mathbf{F}) = Concat(head_1,\dots, head_h)\tilde{\mathbf{W}}^O, \\
% \text{where} \ head_i = Softmax(\frac{\mathbf{Q}_{F,i}\mathbf{K}_i^T}{\sqrt{D^\prime}})\mathbf{V_{F,i}}, \\
% \mathbf{Q}_{F,i} = \mathbf{F}\tilde{\mathbf{W}}^Q_{F,i},\  \mathbf{K}_{F,i} = \mathbf{F}\tilde{\mathbf{W}}^K_{F,i},\  \mathbf{V}_{F,i} = \mathbf{F}\tilde{\mathbf{W}}^V_{F,i},
% \end{gather*}
% where $\tilde{\mathbf{W}}_{F,i}^Q \in \mathbb{R}^{D\times D^\prime}, \tilde{\mathbf{W}}_{F,i}^K \in \mathbb{R}^{D\times D^\prime}, \tilde{\mathbf{W}}_{F,i}^V \in \mathbb{R}^{D\times D^\prime}, \tilde{\mathbf{W}}^O\in\mathbb{R}^{hD^\prime\times D}$. 
The output of the self-attention layer is flattened and fed into an linear layer to produce the scaling weights:
\begin{equation}
    \mathbf{s}=SoftMax(Linear\left(Flatten(MHSA(\mathbf{F}))\right) + \mathbf{w}).
\end{equation}
After undergoing EGD and Online Scaling, the model ultimately outputs an adapted result $\overline{f}=\mathbf{s}^T\mathbf{F}$ that incorporates the optimized predictions from various stages of the process. We denote this model as \modelEGD, whose performance is evaluated is Section~\ref{sec:empirical}. Figure~\ref{fig:forecasting_fix} presents its visualization results, demonstrating that after applying Online Scaling, the prediction outcomes are able to better align with the true values, particularly during periods where the model encounters concept drifts. 

\begin{figure}[!htp]
\centering
\includegraphics[width=0.98\linewidth]{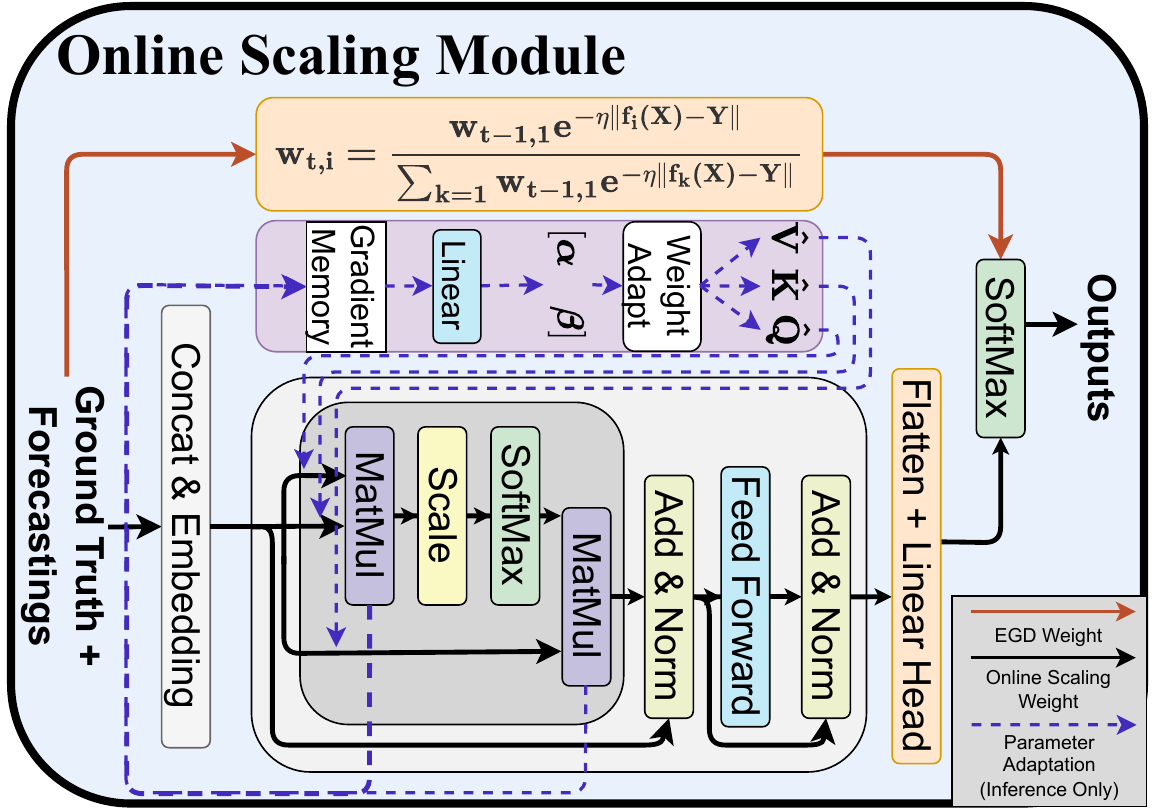}
\caption{The structure of Online Scaling module.}% In the figure, we use solid lines to represent the forward computation process and dashed lines to represent the backward update process.}
\vspace{-4mm}
\label{fig:adapter}
\end{figure}

\subsubsection{Follow the perturbed leader: a more general method}

EGD can achieve guaranteed results, but its effectiveness relies on the assumption that the prediction task can be framed as a convex optimization problem. However, in real-world tasks, it is often difficult to make such an assumption about the environment. Therefore, we also introduce another classical online method, Follow The Perturbed Leader (FTPL)~\cite{kalai2005efficient}, into our algorithm, which we name \modelFTPL.
FTPL has been proven to yield good performance in both convex and non-convex optimization settings~\cite{suggala2020online}. 
Furthermore, since it is essentially a non-parametric approach, it boasts efficiency advantages over EGD, potentially making it a more versatile and generalizable choice.
Specifically, the decision space $\Pi$ encompasses a set of feasible actions or strategies $\mathbf{\pi}_t$ (which expert, or result from a subnetwork, to choose), i.e., $\Pi=\left\{\mathbf{\pi}_t \mid \pi_{t, i}\in \{0,1\} \right.$ and $\left.\left\|\mathbf{\pi}_t\right\|_1=1\right\}$. In an online setting, given $\mathbf{x}$ and $\mathbf{y}$, the algorithm also seeks to minimize the cumulative loss over time.

The core principle of FTPL involves introducing a random perturbation to the decision-making process, thereby allowing for exploration within the decision space. This is formalized as follows:
\begin{equation}\label{eqn:FTPL}
   \mathbf{\pi}_{t+1} = \arg \min_{\mathbf{\pi} \in \Pi} \left\{ \sum_{\tau=1}^t \left\| f_\mathbf{\pi}(\mathbf{X}_\tau)-\mathbf{Y}_\tau\right\|^2 + \sigma(\mathbf{\pi}) \right\}
\end{equation}
where $\sigma(\mathbf{\pi})$ represents the perturbation added to the cumulative loss, often derived from a predefined noise distribution (in this work we use Uniform distribution).
The updating mechanism, according to FTPL, is predicated on combining historical performance data with present stochasticity introduced through $\sigma(\mathbf{\pi})$. This strategic addition of randomness is aimed at mitigating the algorithm's propensity for short-term myopia, thus enhancing its exploratory capabilities and overall adaptability to evolving contexts. The complete procedure of FTPL is shown in Alg.~\ref{alg:FTPL}.

\begin{algorithm}
\DontPrintSemicolon
\KwIn{Pool of different sub-networks $\mathcal{F} = \{f_i\}_{i=1}^d$, Time stamp $t$, Perturbation distribution $\mathbb{D}_{per}$, Loss function $\ell$, Repeat times $m$.}
\KwOut{Target model $f$.}
\For{$j = 1,\dots, m$}{
    \ForEach{base model $f^i \in \mathcal{F}$}
    {Sample the perturbation $\sigma(i)  \sim \mathbb{D}_{per}$;\;}
    Get the decision $\pi_{t+1,j}$ according to the Eqn.~(\ref{eqn:FTPL});\;
}
% Let $\overline{f}_t$ be the the most frequently chosen $f$ during $m$ iterations\;
Sample ${f}$ from the empirical distribution over $\{\pi_{t+1,1},\pi_{t+1,2},\cdots,\pi_{t+1,m}\}$ as the target model;\;
% After the true value $x_t$ is revealed, calculate true loss function $l(\overline{f}(X_t;\theta_t), x_t)$;\;
% Update $\alpha_{t+1}$ according to (\ref{equa:b})\;
% Repeat the choice for next $L$ timestamps and produce related forecasting with confidence interval\;
\caption{Follow The Perturbed Leader (FTPL)}
\label{alg:FTPL}
\end{algorithm}
% In stark contrast to deterministic algorithms, FTPL's central theorem underscores that the introduction of perturbations effectively balances exploitation of known information with the exploration of novel strategies. This balance is crucial for addressing the inherent uncertainty and variability within online environments.
% Crucially, the efficacy and performance of FTPL are directly tied to the nature of the perturbation function $q(\mathbf{\pi})$, which necessitates careful selection to optimize the algorithm's convergence rates and cumulative loss, or regret. 

% \begin{figure}[]
% \centering
% \includegraphics[width=0.98\linewidth]{figure/ensembler.pdf}
% \caption{The pipeline of offline training and online ensemble.}% In the figure, we use solid lines to represent the forward computation process and dashed lines to represent the backward update process.}
% % \vspace{-2mm}
% \label{fig:ensembler}
% \end{figure}

\begin{myTheo}\label{thm:FTPL}\cite{chen2023ipoc}
%Let ${\Pi} \subset \mathbb{R}^d$ be any set (not-necessarily convex) such that $\max _{\mathbf{\pi}, \mathbf{\pi}^{\prime}}\left\|\mathbf{\pi}-\mathbf{\pi}^{\prime}\right\|_1 \leq D$. Assume that the loss functions are such that $\left\|\mathrm{c}^t\right\|_1 \leq A$. Furthermore, assume that for any x and $\mathbf{x},|\mathbf{c} \cdot \mathbf{x}| \leq R$. Then, 
Given decision space $\Pi$, the expectation of regret for Follow the Perturbed Leader
% $q(\pi) \sim %\operatorname{Unif}\left[0, \frac{2}{\epsilon}\right]^n$ has
% \operatorname{Uniform}$ 
has a sublinear bound:
\begin{equation}
   \mathbb{E}[Regret] = \mathbb{E}[\sum_{t=1}^T \mathcal{L}\left(\pi_t\right)-\inf _{\mathbf{u}} \sum_{t=1}^T \mathcal{L}(\mathbf{u})] = O(\sqrt{T})
   %\leq \sqrt{2 A R D T}
    % \leq \frac{2 D}{\epsilon}+T R A \epsilon .
\end{equation}
\end{myTheo}

The theoretical underpinning of FTPL establishes that, under appropriate conditions concerning the perturbation type and algorithmic parameters, it can achieve a vanishing regret that is sublinear in $T$, the total number of time steps, signaling proficient long-term performance. 

We denote this model as \modelFTPL. In the majority of comparisons show in Section~\ref{sec:empirical}, \modelFTPL marginally underperforms relative to \modelEGD. Nonetheless, it offers the benefits of reduced parameter count and expedited inference speeds, contributing to greater throughput as shown in Figure~\ref{fig:throughout} and facilitating flexible deployment in environments limited by resources.

\section{Experiments}\label{sec:empirical}

\subsection{Experimental setting}

\subsubsection{Datasets}

\begin{table}[!htp]
\centering
\caption{Summary of the datasets}
\label{tab:datasets}
\resizebox{\columnwidth}{!}{
\begin{tabular}{cccccc}
\toprule
Category & Dataset     & Domain    & Variates & Samples & Frequency  \\
\midrule
\multirow{6}{*}{Public} & ETTh1       & Industry    & 7        & 17420   & 1 hour \\
&ETTh2       & Industry    & 7        & 17420   & 1 hour \\
&ETTm1       & Industry    & 7        & 69680   & 15 min \\
&ETTm2       & Industry    & 7        & 69680   & 15 min \\
&Electricity & Energy      & 321      & 26304   & 1 hour \\
&Weather     & Environment & 21       & 52696   & 1 hour \\
\midrule
\multirow{6}{*}{Private} &FaaS\_Small     & Private Cloud     & 7        & 23041     & 1 min \\
&FaaS\_Medium     & Private Cloud     & 93        & 23041     & 1 min \\
&FaaS\_Large     & Private Cloud     & 226        & 23041     & 1 min \\
&IaaS\_Small     & Public Cloud     & 7        & 69764     & 1 min \\
&IaaS\_Medium     & Public Cloud     & 58        & 69764     & 1 min \\
&IaaS\_Large     & Public Cloud     & 93        & 69764     & 1 min \\
\bottomrule
\end{tabular}
}
\vspace{-2mm}
\end{table}

% We gather workload data from both public and private cloud systems, and from this collection, we randomly selected different clusters to create datasets of varying sizes. This collection comprise three query-per-second (QPS) datasets {FaaS\_Small, FaaS\_Medium, FaaS\_Large} from private cloud platforms based on Function-as-a-Service (FaaS) and three CPU usage datasets {IaaS\_Small, IaaS\_Medium, IaaS\_Large} from public cloud platforms based on Infrastructure-as-a-Service (IaaS), which exhibit differences in aspects such as periodicity and stationarity due to the platforms they are situated in. We leverage a week's archive of offline data for training the models, which are then deployed in an online configuration. The models forecast workload data at fixed intervals, for instance, every 60 minutes, without any overlap between successive predictions. 
To comprehensively evaluate the performance and generalization capability of our model, we conduct extensive experiments on twelve datasets, including six workload datasets gathered from the Bytedance Cloud platform and six public time series datasets, as shown in Table~\ref{tab:datasets}. Based on the Bytedance Cloud platform, we conduct a 16-day experiment in the private cloud platform and 49-day experiment in the public cloud platform (with workload data at minute-level granularity, predicting once every hour):
\begin{itemize}
    \item The \textbf{FaaS} dataset, including FaaS\_Small, FaaS\_Medium, and FaaS\_Large, comprises query-per-second data gathered from the Function-as-a-Service (FaaS) service.
    \item The \textbf{IaaS} dataset, including IaaS\_Small, IaaS\_Medium, IaaS\_Large, comprises CPU usage data gathered from the Infrastructure-as-a-Service (IaaS) service.
\end{itemize}

Additionally, to assess the generalizability of the proposed model, we carry out extra experiments on public time series datasets:

\begin{itemize}
    \item The \textbf{ETT}\footnotemark[3] dataset, including ETTh1, ETTh2, ETTm1 and ETTm2, comprises load and oil temperature data from two power transformers at different stations.
    \item The \textbf{Electricity}\footnotemark[4] dataset includes electricity consumption data, measured in kilowatt-hours (kWh), from 321 clients, spanning from 2012 to 2014.
    \item The \textbf{Weather}\footnotemark[5] dataset consists of meteorological indicators, sampled every 10 minutes, covering the entire year of 2020.
\end{itemize}

\footnotetext[3]{{\url{https://github.com/zhouhaoyi/ETDataset}}}
\footnotetext[4]{{\url{https://archive.ics.uci.edu/ml/datasets/ElectricityLoadDiagrams20112014}}}
\footnotetext[5]{{\url{https://www.bgc-jena.mpg.de/wetter/}}}

\subsubsection{Baselines}

% We evaluate several baselines for our experiments, including methods for time series forecasting, and online learning. 
We selected four categories of time series forecasting models as baselines:
\begin{enumerate}
    \item Extensively applied traditional statistical models such as ETS~\cite{holt2004forecasting}, Seasonal ARIMA~\cite{williams2003modeling}, and STL~\cite{cleveland1990stl}.
    \item Representative deep neural networks for time series forecasting models (tested with retraining setting), such as MLP-based DLinear~\cite{zeng2023transformers}, CNN-based TimesNet~\cite{wutimesnet}, Transformer-based iTransformer~\cite{liuitransformer} and PatchTST~\cite{nietime}.
    \item LLM-based time series forecasting models designed for few-shot and zero-shot forecasting, GPT4TS~\cite{zhou2023one} (tested in two manners: offline forecasting following the original setting and online forecasting with retraining).
    \item Neural networks for online prediction, such as FSNet~\cite{phamlearning}, OneNet~\cite{wen2024onenet}, and Time-FSNet, a variant of FSNet, which modifies the dimensionality of convolutional computations.
    
% FSNet~\cite{phamlearning} is a typical model for online time series forecasting, which features a convolutional layer with an Online Adaptor, enabling online adaptive updates of model parameters. Time-FSNet, a variant of FSNet, modifies the dimensionality of convolutional computations, which shifts the process from convolving across the feature dimension to convolving across the time dimension. OneNet~\cite{wen2024onenet}, an ensemble model that includes both an FSNet and an Time-FSNet, obtaining the final result through weighted summation, is the previous state-of-the-art online forecasting method. In addition to online models, to ensure the diversity of model structures, we compared several advanced time series prediction models using an online retraining strategy.
\end{enumerate}
% The former primarily explores the correlations between variables, while the latter adopts a channel-independence setting.
% TimesNet~\cite{wutimesnet} is a convolution-based time series model.
% transforms 1-dimensional time series into 2-dimensional representations for computation using 2D convolution. 
% DLinear~\cite{zeng2023transformers} is is a simple but effective forecasting model includes sequence decomposition and MLP layers.

\begin{table*}[!htbp]
  \caption{Results for the forecasting task on the cloud system workload datasets. The input sequence length is set to $1440$. All the results are averaged from 4 different forecasting horizons $\{1, 10, 30, 60\}$. }
  \vspace{-2mm}
  \label{tab:forecasting_results_1}
  \vskip 0.05in
  \centering
  \resizebox{\linewidth}{!}{
  \begin{threeparttable}
  \begin{small}
  \renewcommand{\multirowsetup}{\centering}
  \setlength{\tabcolsep}{1pt}
  \begin{tabular}{cc|>{\centering\arraybackslash}p{0.8cm}>{\centering\arraybackslash}p{0.8cm}>{\centering\arraybackslash}p{0.8cm}|>{\centering\arraybackslash}p{0.8cm}>{\centering\arraybackslash}p{0.8cm}>{\centering\arraybackslash}p{0.8cm}|>{\centering\arraybackslash}p{0.8cm}>{\centering\arraybackslash}p{0.8cm}>{\centering\arraybackslash}p{0.8cm}|>{\centering\arraybackslash}p{0.8cm}>{\centering\arraybackslash}p{0.8cm}>{\centering\arraybackslash}p{0.8cm}|>{\centering\arraybackslash}p{0.8cm}>{\centering\arraybackslash}p{0.8cm}>{\centering\arraybackslash}p{0.8cm}|>{\centering\arraybackslash}p{0.8cm}>{\centering\arraybackslash}p{0.8cm}>{\centering\arraybackslash}p{0.8cm}}
    \toprule
    \multicolumn{2}{c|}{Datasets} & 
    \multicolumn{3}{c}{\rotatebox{0}{{FaaS\_Small}}} &
    \multicolumn{3}{c}{\rotatebox{0}{{FaaS\_Medium}}} &
    \multicolumn{3}{c|}{\rotatebox{0}{{{FaaS\_Large}}}} &
    \multicolumn{3}{c}{\rotatebox{0}{{IaaS\_Small}}} &
    \multicolumn{3}{c}{\rotatebox{0}{{IaaS\_Medium}}} &
    \multicolumn{3}{c}{\rotatebox{0}{{IaaS\_Lage}}}\\

    % \multicolumn{2}{c}{} & \multicolumn{3}{c}{{(cite)}} & 
    % \multicolumn{3}{c|}{{(cite)}} &
    % \multicolumn{3}{c}{{\cite{wen2024onenet}}} &
    % \multicolumn{3}{c}{{\cite{phamlearning}}} & 
    % \multicolumn{3}{c|}{{\cite{phamlearning}}} & 
    % \multicolumn{3}{c}{{\cite{liuitransformer}}} & 
    % \multicolumn{3}{c}{{\cite{nietime}}} &
    % \multicolumn{3}{c}{{\cite{zeng2023transformers}}} & 
    % \multicolumn{3}{c}{{\cite{wutimesnet}}}\\
    
    % \cmidrule(lr){3-5} \cmidrule(lr){6-8}\cmidrule(lr){9-11} \cmidrule(lr){12-14}\cmidrule(lr){15-17}\cmidrule(lr){18-20}\cmidrule(lr){21-23}\cmidrule(lr){24-26} \cmidrule(lr){27-29}
    \multicolumn{2}{c|}{Metric} & \centering\arraybackslash\scalebox{0.7}{MSE} & \centering\arraybackslash\scalebox{0.7}{MAE} & \centering\arraybackslash\scalebox{0.7}{\underline{WMAPE}} & \centering\arraybackslash\scalebox{0.7}{MSE} & \centering\arraybackslash\scalebox{0.7}{MAE} & \centering\arraybackslash\scalebox{0.7}{\underline{WMAPE}} & \centering\arraybackslash\scalebox{0.7}{MSE} & \centering\arraybackslash\scalebox{0.7}{MAE} & \centering\arraybackslash\scalebox{0.7}{\underline{WMAPE}} & \centering\arraybackslash\scalebox{0.7}{MSE} & \centering\arraybackslash\scalebox{0.7}{MAE} & \centering\arraybackslash\scalebox{0.7}{WMAPE} & \centering\arraybackslash\scalebox{0.7}{MSE} & \centering\arraybackslash\scalebox{0.7}{MAE} & \centering\arraybackslash\scalebox{0.7}{WMAPE} & \centering\arraybackslash\scalebox{0.7}{MSE} & \centering\arraybackslash\scalebox{0.7}{MAE} & \centering\arraybackslash\scalebox{0.7}{WMAPE}\\
    \toprule
    {\rotatebox{0}
    {\scalebox{0.95}STL\cite{cleveland1990stl}}}
    &  & {1.303}&{0.803}&{6.339}&{1.453}&{0.860}&{6.885}&{3.168}&{1.105}&{21.320}&{1.750}&{1.006}&{1.317}&{1.715}&{1.034}&{1.225}&{1.742}&{1.032}&{1.131}\\
    {\rotatebox{0}
    {\scalebox{0.95}SARIMA\cite{williams2003modeling}}}
    &  & {1.150}&{0.728}&{4.102}&{1.288}&{0.790}&{6.511}&{3.025}&{1.059}&{20.685}&{1.237}&{0.836}&{1.065}&{1.290}&{0.891}&{1.034}&{1.329}&{0.888}&{0.958}\\
    {\rotatebox{0}
    {\scalebox{0.95}ETS\cite{holt2004forecasting}}}
    &  & {1.222}&{0.749}&{6.113}&{1.358}&{0.811}&{6.662}&{3.128}&{1.073}&{20.915}&{1.318}&{0.858}&{1.097}&{1.305}&{0.896}&{1.038}&{1.337}&{0.887}&{0.952}\\
    \midrule
    {\rotatebox{0}
    {\scalebox{0.95}DLinear\cite{zeng2023transformers}}}
    &  &3.401   & 1.003 &  7.225 &  3.368 & 1.033 & 10.602  & 0.267  & 0.288 & 3.629  & 0.748  & 0.615 & 0.737  & 0.902  &0.753  & 0.827  & 0.970  &0.773  &0.816  \\
    {\rotatebox{0}{\scalebox{0.95}TimesNet\cite{wutimesnet}}}
    &  &  0.596 & 0.542 & 3.846 & 0.525 & 0.499 & 3.423 & 0.922 & 0.571 & 10.015 & 0.828 & 0.665 & 0.829 &0.858  &0.769  &0.862  &0.901  &0.794  &0.828  \\
    {\rotatebox{0}{\scalebox{0.95}PatchTST\cite{nietime}}}
    &  &0.327 & 0.370 & 3.787 & 0.290 & 0.334 & 1.999 & 0.479 & 0.391 & 6.728 & 0.689  & 0.582  & 0.701  & 0.804  & 0.711  & 0.787 & 0.830 &0.723  & 0.756 \\
    {\rotatebox{0}{\scalebox{0.95}iTransformer\cite{liuitransformer}}}
    &  &0.298   &0.352  & 2.345  & 0.287  &0.331  &1.977   & 0.340  &0.318  &5.171   &0.840   & 0.692 & 0.838  &  0.830 &0.753  & 0.831  & 0.868  & 0.759 & 0.775\\
    \midrule
    {\rotatebox{0}{\scalebox{0.95}GPT4TS\cite{zhou2023one}}}
    & & {0.262}&{0.333}&{2.145}&{0.235}&{0.296}&{1.715}&{0.245}&\boldres{0.267}&{4.042}&{0.666}&\secondres{0.565}&{0.679}&{0.755}&\boldres{0.675}&\boldres{0.743}&\secondres{0.749}&{0.701}&{0.718}\\
    {\rotatebox{0}{\scalebox{0.95}Online-GPT4TS}}
    & &{0.253}&{0.329}&{2.096}&{0.244}&{0.314}&{1.927}&{0.286}&{0.318}&{5.343}&\secondres{0.645}&{0.577}&{0.717}&{0.760}&{0.700}&{0.767}&{0.765}&{0.719}&{0.733}\\
    % {\rotatebox{0}{\scalebox{0.95}Uni2TS\cite{woounified}}}
    % \\
    \midrule
    {\rotatebox{0}{\scalebox{0.95}FSNet\cite{phamlearning}}}
    &  &  0.271 & 0.321 & 1.930 & 0.265 & 0.325 & 2.281 & 0.266 & 0.329 & 2.425 & 0.672 & 0.588 & 0.696 & 0.797 & 0.721 & 0.783 & 0.841 & 0.743 & 0.776\\
    {\rotatebox{0}{\scalebox{0.95}Time-FSNet\cite{phamlearning}}}
    &  &0.249 & 0.320 & 2.054 & 0.228 & 0.298 & 1.827 & 0.237 & 0.306 & 1.801 & 0.734 & 0.629 & 0.754 & 0.790 & 0.732 & 0.807 & 0.833 & 0.746 & 0.775 \\
    {\rotatebox{0}{\scalebox{0.95}OneNet\cite{wen2024onenet}}}
    &  & 0.228 & 0.294 & 1.709 & 0.229 & 0.293 & 1.781 & 0.229 & 0.293 & 1.781 & 0.674 & 0.599 & 0.705& 0.777 & 0.718 & 0.785 & 0.761 & 0.709 & 0.725 \\
    \midrule
    {\rotatebox{0}{\textbf{\scalebox{0.95}\modelFTPL}}}
    & &\secondres{0.208} & \secondres{0.270} & \secondres{1.477} & \secondres{0.214} & \secondres{0.274} & \secondres{1.381} & \boldres{0.214} & \secondres{0.274} & \boldres{1.385} & {0.646} & \secondres{0.565} & \secondres{0.669} & \secondres{0.741} & {0.692} & {0.751} & {0.753} & \secondres{0.691} & \secondres{0.707} \\
    {\rotatebox{0}{\textbf{\scalebox{0.95}\modelEGD}}}
    &  &\boldres{0.203} & \boldres{0.263} & \boldres{1.419} & \boldres{0.208} & \boldres{0.267} & \boldres{1.281} & \secondres{0.217} & {0.277} & \secondres{1.445} & \boldres{0.634} & \boldres{0.563} & \boldres{0.665} & \boldres{0.733} & \secondres{0.689} & \secondres{0.746} & \boldres{0.734} & \boldres{0.682} & \boldres{0.695}\\
    \bottomrule
  \end{tabular}
    \begin{tablenotes}
        \footnotesize
        \item[] \underline{WMAPE} indicates that we multiply the WMAPE by 100 on FaaS data to better display and compare the results.
  \end{tablenotes}
    \end{small}
  \end{threeparttable}
  }
\end{table*}

\subsubsection{Implemental Details}
For FaaS and IaaS datasets, we leverage a week's archive of offline data for training the models, which are deployed in an online configuration. The models forecast workload data at fixed intervals, for instance, every 60 minutes, without any overlap between successive predictions to conserve computational resources. Over the course of our research, the model performs online predictions for a duration of one week on FaaS and six weeks on IaaS. We follow the optimization details in Informer~\cite{zhou2023expanding} by optimizing the $\ell_2$ (MSE) loss with the AdamW optimizer~\cite{loshchilovdecoupled}. For all benchmarks, we set the look-back window length to be $1440$ (is the maximum reasonable query length that the cloud computing system we are using can handle online) and vary the forecast horizon as $H\in\{1, 10, 30, 60 \}$. During the training phase, the batch size for all models is set to $32$, except for GPT4TS, which is set to $8$ to avoid GPU memory overflow. The hyperparameters for all models are configured based on their official implementations or implementations found in other research papers. Specifically, for the proposed model, we chose a group of patch sizes $\{16, 32, 64, 128\}$ and During the parameter analysis, this selection will be expanded to include $\{16, 32, 64, 128, 256, 512\}$.

All experiments are repeated three times, implemented in PyTorch and conducted on NVIDIA Tesla V100 32GB GPUs. All the baselines are implemented based on configurations of the original paper or official code. For the metrics, we adopt the mean square error (MSE), mean absolute error (MAE), and mean absolute percentage errors (WMAPE) for long-term forecasting and imputations, where MSE and MAE are calculated after the data has been standard normalized, whereas WMAPE is calculated after normalization and denormalization. These metrics can be calculated as follows:
$$
MSE =
% \mathcal{L}=\frac{1}{M} \sum_{j=1}^M\left\|\hat{x}_{K^{\prime}+1: K^{\prime}+H}^j-x_{K^{\prime}+1: K^{\prime}+H}^j\right\| .
\frac{1}{M} \sum_{j=1}^M\left\|\hat{x}_{t+1: t+H}^j-x_{t+1: t+H}^j\right\|,
$$
$$
MAE =
% \mathcal{L}=\frac{1}{M} \sum_{j=1}^M\left\|\hat{x}_{K^{\prime}+1: K^{\prime}+H}^j-x_{K^{\prime}+1: K^{\prime}+H}^j\right\| .
\frac{1}{M} \sum_{j=1}^M\left|\hat{x}_{t+1: t+H}^j-x_{t+1: t+H}^j\right|,
$$
$$
WMAPE =
% \mathcal{L}=\frac{1}{M} \sum_{j=1}^M\left\|\hat{x}_{K^{\prime}+1: K^{\prime}+H}^j-x_{K^{\prime}+1: K^{\prime}+H}^j\right\| .
\frac{\sum_{j=1}^M\left|\hat{x}_{t+1: t+H}^j-x_{t+1: t+H}^j\right|}{\sum_{k=1}^M x_{t+1: t+H}^k}.
$$

\subsection{Online Forecasting Results}
In this setting, the data used for training and online forecasting is from the same time series.
Table~\ref{tab:forecasting_results_1} and Table~\ref{tab:forecasting_results_2} present the cumulative performance of different baselines in terms of mean-squared errors (MSE), mean-absolute errors (MAE), and weighted mean absolute percentage errors (WMAPE), where MSE and MAE are calculated after the data has been standard normalized, whereas WMAPE is calculated after normalization and denormalization. The workload data from FaaS demonstrates significant periodicity, typically leading to comparatively lower prediction errors in models. Models operating online, like OneNet, generally surpass the performance of offline models that necessitate retraining, exemplified by iTransformer. This performance discrepancy is attributed to the clear data drift characteristic of FaaS data. Despite the data being minutely granular, it also captures broader cycles on daily, weekly, and monthly scales. Offline models, despite undergoing retraining, face difficulties in accurately adapting to these evolving features. LLM-based GPT4TS shows good performance, but its parameter volume far exceeds that of other models. Our proposed \modelEGD and \modelFTPL have demonstrated superior performance compared to the baselines. Specifically, \modelEGD reduces MSE by 13.9\%, MAE by 11.7\%, and WMAPE by 19.3\% on average across seven datasets compared to the best baseline, OneNet (\modelFTPL exceeded it by 9.2\%/8.5\%/13.3\%).

\begin{table*}[!htbp]
  \caption{Results for the forecasting task on the other public datasets. The input sequence length is set to $1440$. All the results are averaged from 4 different forecasting horizons $\{1, 10, 30, 60\}$. }%See Table~\ref{tab:full_forecasting_results} in Appendix for the full results. }%\xiao{It is difficult to find the improvement of our method compared with the others. Maybe use one metric in a single table? Or each row represents results for a method?}}
  \vspace{-2mm}
  \label{tab:forecasting_results_2}
  \vskip 0.05in
  \centering
  \resizebox{\linewidth}{!}{
  \begin{threeparttable}
  \begin{small}
  \renewcommand{\multirowsetup}{\centering}
  \setlength{\tabcolsep}{1pt}
  \begin{tabular}{cc|>{\centering\arraybackslash}p{0.8cm}>{\centering\arraybackslash}p{0.8cm}>{\centering\arraybackslash}p{0.8cm}|>{\centering\arraybackslash}p{0.8cm}>{\centering\arraybackslash}p{0.8cm}>{\centering\arraybackslash}p{0.8cm}|>{\centering\arraybackslash}p{0.8cm}>{\centering\arraybackslash}p{0.8cm}>{\centering\arraybackslash}p{0.8cm}|>{\centering\arraybackslash}p{0.8cm}>{\centering\arraybackslash}p{0.8cm}>{\centering\arraybackslash}p{0.8cm}|>{\centering\arraybackslash}p{0.8cm}>{\centering\arraybackslash}p{0.8cm}>{\centering\arraybackslash}p{0.8cm}|>{\centering\arraybackslash}p{0.8cm}>{\centering\arraybackslash}p{0.8cm}>{\centering\arraybackslash}p{0.8cm}}
    \toprule
    \multicolumn{2}{c|}{Datasets} & 
    \multicolumn{3}{c|}{\rotatebox{0}{ETTh1}} &
    \multicolumn{3}{c|}{\rotatebox{0}{ETTh2}} &
    \multicolumn{3}{c|}{\rotatebox{0}{ETTm1}} &
    \multicolumn{3}{c|}{\rotatebox{0}{ETTm2}} &
    \multicolumn{3}{c|}{\rotatebox{0}{ECL}} &
    \multicolumn{3}{c}{\rotatebox{0}{Weather}}\\

    % \multicolumn{2}{c}{} & \multicolumn{3}{c}{{(cite)}} & 
    % \multicolumn{3}{c|}{{(cite)}} &
    % \multicolumn{3}{c}{{\cite{wen2024onenet}}} &
    % \multicolumn{3}{c}{{\cite{phamlearning}}} & 
    % \multicolumn{3}{c|}{{\cite{phamlearning}}} & 
    % \multicolumn{3}{c}{{\cite{liuitransformer}}} & 
    % \multicolumn{3}{c}{{\cite{nietime}}} &
    % \multicolumn{3}{c}{{\cite{zeng2023transformers}}} & 
    % \multicolumn{3}{c}{{\cite{wutimesnet}}}\\
    
    % \cmidrule(lr){3-5} \cmidrule(lr){6-8}\cmidrule(lr){9-11} \cmidrule(lr){12-14}\cmidrule(lr){15-17}\cmidrule(lr){18-20}\cmidrule(lr){21-23}\cmidrule(lr){24-26} \cmidrule(lr){27-29}
    \multicolumn{2}{c|}{Metric} & \centering\arraybackslash\scalebox{0.7}{MSE} & \centering\arraybackslash\scalebox{0.7}{MAE} & \centering\arraybackslash\scalebox{0.7}{{WMAPE}} & \centering\arraybackslash\scalebox{0.7}{MSE} & \centering\arraybackslash\scalebox{0.7}{MAE} & \centering\arraybackslash\scalebox{0.7}{{WMAPE}} & \centering\arraybackslash\scalebox{0.7}{MSE} & \centering\arraybackslash\scalebox{0.7}{MAE} & \centering\arraybackslash\scalebox{0.7}{{WMAPE}} & \centering\arraybackslash\scalebox{0.7}{MSE} & \centering\arraybackslash\scalebox{0.7}{MAE} & \centering\arraybackslash\scalebox{0.7}{WMAPE} & \centering\arraybackslash\scalebox{0.7}{MSE} & \centering\arraybackslash\scalebox{0.7}{MAE} & \centering\arraybackslash\scalebox{0.7}{WMAPE} & \centering\arraybackslash\scalebox{0.7}{MSE} & \centering\arraybackslash\scalebox{0.7}{MAE} & \centering\arraybackslash\scalebox{0.7}{WMAPE}\\
    \toprule
    {\rotatebox{0}
    {\scalebox{0.95}STL\cite{cleveland1990stl}}}
    &  & {1.757} & {0.992} & {1.176} & {1.702} & {0.933} & {0.562} & {3.152} & {1.283} & {0.756} & {3.938} & {1.368} & {0.323} & {1.484} & {0.821} & {0.335} & {6.555} & {1.697} & {0.576}\\
    {\rotatebox{0}
    {\scalebox{0.95}SARIMA\cite{williams2003modeling}}}
    &  & {1.890} & {0.917} & {1.021} & {1.689} & {0.905} & {0.541} & {2.882} & {1.137} & {0.671} & {3.688} & {1.239} & {0.292} & {1.337} & {0.776} & {0.316} & {5.594} & {1.441} & {0.475}\\
    {\rotatebox{0}
    {\scalebox{0.95}ETS\cite{holt2004forecasting}}}
    &  & {2.563} & {1.062} & {1.166} & {1.654} & {0.897} & {0.537} & {2.899} & {1.145} & {0.675} & {3.753} & {1.255} & {0.294} & {1.511} & {0.807} & {0.328} & {5.820} & {1.468} & {0.485}\\
    \midrule
    {\rotatebox{0}
    {\scalebox{0.95}DLinear\cite{zeng2023transformers}}}
    &  &\secondres{0.282} & \secondres{0.359} & {0.376} & {0.162} & {0.259} & {0.149} & {0.346} & {0.429} & {0.219} & {0.355} & {0.391} & {0.073} & {0.182} & {0.254} & {0.090} & {0.642} & {0.374} & {0.107}\\
    {\rotatebox{0}{\scalebox{0.95}TimesNet\cite{wutimesnet}}}
    &  &  {0.464} & {0.482} & {0.521} & {0.307} & {0.386} & {0.225} & {0.671} & {0.561} & {0.301} & {0.621} & {0.558} & {0.119} & {0.352} & {0.348} & {0.112} & {1.457} & {0.754} & {0.165}  \\
    {\rotatebox{0}{\scalebox{0.95}PatchTST\cite{nietime}}}
    &  &{0.335} & {0.401} & {0.419} & {0.160} & {0.268} & {0.157} & {0.398} & {0.414} & {0.214} & {0.366} & {0.400} & {0.077} & {0.201} & {0.257} & {0.083} & {0.815} & {0.458} & {0.124}\\
    {\rotatebox{0}{\scalebox{0.95}iTransformer\cite{liuitransformer}}}
    &  &{0.294} & {0.371} & {0.401} & {0.180} & {0.281} & {0.163} & {0.503} & {0.485} & {0.263} & {0.520} & {0.510} & {0.106} & {0.202} & {0.258} & {0.088} & {1.057} & {0.605} & {0.154}\\
    \midrule
    {\rotatebox{0}{\scalebox{0.95}GPT4TS\cite{zhou2023one}}}
    & & {0.311} & {0.383} & {0.419} & {0.160} & {0.264} & {0.153} & \secondres{0.319} & {0.385} & {0.205} & {0.351} & {0.410} & {0.084} &- &- & -&0.688  &0.419  &0.109 \\
    {\rotatebox{0}{\scalebox{0.95}Online-GPT4TS}}
    & & {0.382} & {0.430} & {0.482} & {0.174} & {0.277} & {0.159} & {0.405} & {0.451} & {0.248} & {0.434} & {0.463} & {0.095}&- &- & -&1.195  &0.658  &0.183 \\
    % {\rotatebox{0}{\scalebox{0.95}Uni2TS\cite{woounified}}}
    % \\
    \midrule
    {\rotatebox{0}{\scalebox{0.95}FSNet\cite{phamlearning}}}
    &  &  {0.354} & {0.416} & {0.418} & {0.213} & {0.311} & {0.178} & {0.651} & {0.534} & {0.290} & {1.825} & {0.903} & {0.205} & {0.310} & {0.349} & {0.121} & {0.952} & {0.598} & {0.220}\\
    {\rotatebox{0}{\scalebox{0.95}Time-FSNet\cite{phamlearning}}}
    &  &{0.315} & {0.389} & {0.418} & {0.267} & {0.356} & {0.201} & {0.443} & {0.479} & {0.284} & {0.431} & {0.460} & {0.096} & {0.198} & {0.280} & {0.096} & {0.959} & {0.590} & {0.184}\\
    {\rotatebox{0}{\scalebox{0.95}OneNet\cite{wen2024onenet}}}
    &  & {0.297} & {0.374} & {0.387} & {0.188} & {0.296} & {0.166} & {0.395} & {0.432} & {0.241} & {0.417} & {0.444} & {0.090} & {0.189} & {0.272} & {0.095} & {0.694} & {0.430} & {0.135}\\
    \midrule
    {\rotatebox{0}{\textbf{\scalebox{0.95}\modelFTPL}}}
    & &{0.285} & {0.362} & \secondres{0.373} & \secondres{0.155} & \secondres{0.258} & \secondres{0.148} & \secondres{0.319} & \secondres{0.384} & \secondres{0.195} & \secondres{0.333} & \secondres{0.374} & \secondres{0.071} & \secondres{0.180} & \secondres{0.246} & \secondres{0.087} & \boldres{0.602} & \boldres{0.356} & \boldres{0.099} \\
    {\rotatebox{0}{\textbf{\scalebox{0.95}\modelEGD}}}
    &  &\boldres{0.275} & \boldres{0.352} & \boldres{0.365} & \boldres{0.150} & \boldres{0.252} & \boldres{0.143} & \boldres{0.282} & \boldres{0.342} & \boldres{0.174} & \boldres{0.328} & \boldres{0.371} & \boldres{0.071} & \boldres{0.171} & \boldres{0.242} & \boldres{0.083} & \secondres{0.626} & \secondres{0.368} & \secondres{0.103}\\
    \bottomrule
  \end{tabular}
      \begin{tablenotes}
        \footnotesize
        \item[] - CUDA out of memory.
  \end{tablenotes}
    \end{small}
  \end{threeparttable}
  }
\end{table*}

\begin{table*}[!htbp]
  \caption{Results for the transfer forecasting task. All models are pre-trained on the source dataset and tested on the target dataset. For the same domain, such as FaaS, we randomly collect 7 different clusters from the system to construct the souce and target. All the results are averaged from 4 different
forecasting horizons \{1, 10, 30, 60\}. }
  \vspace{-2mm}
  \label{tab:transfer_results}
  \vskip 0.05in
  \centering
  \resizebox{\linewidth}{!}{
  \begin{threeparttable}
  \begin{small}
  \renewcommand{\multirowsetup}{\centering}
  \setlength{\tabcolsep}{1pt}
  \begin{tabular}{cc|>{\centering\arraybackslash}p{0.8cm}>{\centering\arraybackslash}p{0.8cm}>{\centering\arraybackslash}p{0.8cm}|>{\centering\arraybackslash}p{0.8cm}>{\centering\arraybackslash}p{0.8cm}>{\centering\arraybackslash}p{0.8cm}|>{\centering\arraybackslash}p{0.8cm}>{\centering\arraybackslash}p{0.8cm}>{\centering\arraybackslash}p{0.8cm}|>{\centering\arraybackslash}p{0.8cm}>{\centering\arraybackslash}p{0.8cm}>{\centering\arraybackslash}p{0.8cm}|>{\centering\arraybackslash}p{0.8cm}>{\centering\arraybackslash}p{0.8cm}>{\centering\arraybackslash}p{0.8cm}|>{\centering\arraybackslash}p{0.8cm}>{\centering\arraybackslash}p{0.8cm}>{\centering\arraybackslash}p{0.8cm}|>{\centering\arraybackslash}p{0.8cm}>{\centering\arraybackslash}p{0.8cm}>{\centering\arraybackslash}p{0.8cm}}
    \toprule
    \multicolumn{2}{c|}{Datasets} & 
    \multicolumn{3}{c}{\rotatebox{0}{{FaaS$\rightarrow$FaaS}}} &
    \multicolumn{3}{c|}{\rotatebox{0}{{IaaS$\rightarrow$IaaS}}} &
    \multicolumn{3}{c}{\rotatebox{0}{{{FaaS$\rightarrow$IaaS}}}} &
    \multicolumn{3}{c|}{\rotatebox{0}{{IaaS$\rightarrow$FaaS}}} &
    \multicolumn{3}{c}{\rotatebox{0}{{ETTh1$\rightarrow$ETTm1}}} &
    \multicolumn{3}{c|}{\rotatebox{0}{\scalebox{0.9}{ETTm2$\rightarrow$ETTh2}}} &
    \multicolumn{3}{c}{\rotatebox{0}{\scalebox{0.9}{FaaS$\rightarrow$Weather}}} \\

    \multicolumn{2}{c|}{Metric} & \centering\arraybackslash\scalebox{0.7}{MSE} & \centering\arraybackslash\scalebox{0.7}{MAE} & \centering\arraybackslash\scalebox{0.7}{WMAPE} &
    \centering\arraybackslash\scalebox{0.7}{MSE} & \centering\arraybackslash\scalebox{0.7}{MAE} & \centering\arraybackslash\scalebox{0.7}{WMAPE} &
    \centering\arraybackslash\scalebox{0.7}{MSE} & \centering\arraybackslash\scalebox{0.7}{MAE} & \centering\arraybackslash\scalebox{0.7}{WMAPE} &
    \centering\arraybackslash\scalebox{0.7}{MSE} & \centering\arraybackslash\scalebox{0.7}{MAE} & \centering\arraybackslash\scalebox{0.7}{WMAPE} &
    \centering\arraybackslash\scalebox{0.7}{MSE} & \centering\arraybackslash\scalebox{0.7}{MAE} & \centering\arraybackslash\scalebox{0.7}{WMAPE} &
    \centering\arraybackslash\scalebox{0.7}{MSE} & \centering\arraybackslash\scalebox{0.7}{MAE} & \centering\arraybackslash\scalebox{0.7}{WMAPE} &
    \centering\arraybackslash\scalebox{0.7}{MSE} & \centering\arraybackslash\scalebox{0.7}{MAE} & \centering\arraybackslash\scalebox{0.7}{WMAPE} \\
    \toprule
    {\rotatebox{0}
    {\scalebox{0.95}STL\cite{cleveland1990stl}}}
    &  & {4.851}&{1.228}&{0.196}&{1.717}&{1.024}&{1.007}&{1.750}&{1.006}&{1.317}&{1.303}&{0.803}&{0.063}&{3.152}&{1.283}&{0.756}&{1.423}&{0.773}&{0.351}&6.555&1.697&0.576\\
    {\rotatebox{0}
    {\scalebox{0.95}SARIMA\cite{williams2003modeling}}}
    &  &{4.726}&{1.193}&{0.194}&{1.353}&{0.889}&{0.865}&{1.237}&{0.836}&{1.065}&{1.150}&{0.728}&{0.059}&{2.882}&{1.137}&{0.671}&{1.297}&{0.710}&{0.323}&5.594  &1.441  &0.475  \\
    {\rotatebox{0}
    {\scalebox{0.95}ETS\cite{holt2004forecasting}}}
    &  & {4.696} & {1.187} & {0.191} & {1.424} & {0.923} & {0.900} & {1.339} & {0.886} & {1.131} & {1.261} & {0.761} & {0.061} & {2.899} & {1.145} & {0.675} & {1.290} & {0.712} & {0.324}&5.542  &1.441  &0.476 \\
    \midrule
    {\rotatebox{0}{\scalebox{0.95}DLinear\cite{zeng2023transformers}}}&&12.478&1.785&0.389&1.339&0.947&0.966&1.363&0.900&1.230&0.661&0.530&0.037&0.417&0.442&0.226&\secondres{0.275}&0.303&\secondres{0.119}& 2.514 & 0.897 &0.329 \\
    {\rotatebox{0}{\scalebox{0.95}TimesNet\cite{wutimesnet}}}&&2.233&0.938&0.161&1.451&0.987&0.988&1.408&0.926&1.151&0.901&0.695&0.057&0.769&0.644&0.347&0.752&0.528&0.229 &5.053  &1.521  &0.470 \\
    {\rotatebox{0}{\scalebox{0.95}PatchTST\cite{nietime}}}&&1.027&0.571&0.087&1.165&0.888&0.876&0.999&0.785&0.930&3.104&0.845&0.050&0.550&0.498&0.256&0.320&0.341&0.138&2.135  &0.892  &0.302 \\
    {\rotatebox{0}{\scalebox{0.95}iTransformer\cite{liuitransformer}}}&&0.714&0.431&0.061&1.133&0.871&0.848&1.306&0.887&1.079&\secondres{0.617}&0.567&0.041&0.718&0.610&0.335&0.333&0.355&0.148 &1.698  & 0.736 &0.215 \\
    \midrule
    {\rotatebox{0}{\scalebox{0.95}GPT4TS\cite{zhou2023one}}}
    & &{0.592}&{0.380}&{0.049}&{1.003}&{0.856}&{0.834}&{1.159}&{0.810}&{0.986}&{0.870}&{0.657}&{0.047}&{0.461}&{0.463}&{0.249}&{0.305}&{0.329}&{0.136}&1.411 &0.681 &0.208\\
    {\rotatebox{0}{\scalebox{0.95}Online-GPT4TS}}
    & &{0.718}&{0.508}&{0.086}&\secondres{0.993}&\secondres{0.842}&\secondres{0.827}&{1.090}&{0.814}&{0.998}&{0.623}&\secondres{0.482}&\secondres{0.033}&{0.448}&{0.475}&{0.256}&{0.338}&{0.346}&{0.143}&1.202  &0.670  & 0.188 \\
    % {\rotatebox{0}{\scalebox{0.95}Uni2TS\cite{woounified}}}
    % \\
    \midrule
    {\rotatebox{0}{\scalebox{0.95}FSNet\cite{phamlearning}}}& &1.227&0.566&0.081&1.309&0.940&0.973&1.398&0.955&1.221&1.517&0.913&0.077&0.603&0.553&0.318&0.670&0.544&0.228& 1.539 &0.810  &0.286 \\
    {\rotatebox{0}{\scalebox{0.95}Time-FSNet}}& &0.949&0.586&0.100&1.107&0.876&{0.846}&1.016&0.811&0.955&0.594&0.545&0.041&0.556&0.312&0.533&0.303&0.356&0.149& 0.966  &0.623  &0.236 
    \\
    {\rotatebox{0}{\scalebox{0.95}OneNet\cite{wen2024onenet}}}& &0.511&0.441&0.077&{1.082}&0.868&0.871&\secondres{0.965}&\secondres{0.771}&\secondres{0.916}&0.627&0.559&0.039&0.424&0.456&0.252&0.323&0.359&0.152& \secondres{1.052} &\secondres{0.609}  &0.185 
    \\
    \midrule
    {\rotatebox{0}{\textbf{\scalebox{0.95}\modelFTPL}}}& &\secondres{0.499}&\secondres{0.345}&\secondres{0.045}&1.086&{0.864}&0.860&0.970&0.783&0.924&0.623&{0.501}&{0.034}&\secondres{0.389}&\secondres{0.419}&\secondres{0.222}&\secondres{0.275}&\secondres{0.302}&0.120&1.086  &0.610  &\secondres{0.179 }
    \\
    {\rotatebox{0}{\textbf{\scalebox{0.95}\modelEGD}}}&&\boldres{\centering 0.396} & \boldres{\centering 0.330} & \boldres{\centering 0.042} & \boldres{\centering 0.982} & \boldres{\centering 0.829} & \boldres{\centering 0.823} &\boldres{\centering 0.909} & \boldres{\centering 0.749} & \boldres{\centering 0.885} &\boldres{\centering 0.531} & \boldres{\centering 0.469} & \boldres{\centering 0.031} &\boldres{\centering 0.356} & \boldres{\centering 0.405} & \boldres{\centering 0.215} & \boldres{\centering 0.264} & \boldres{\centering 0.296} & \boldres{\centering 0.117}&\boldres{1.029}  &\boldres{0.581}  &\boldres{0.159} \\
    \bottomrule
  \end{tabular}
  %   \begin{tablenotes}
  %       \footnotesize
  %       \item[] $\ast$ means that there are some mismatches between our input-output setting and their papers. We adopt their official codes and only change the length of input and output sequences for a fair comparison.
  % \end{tablenotes}
    \end{small}
  \end{threeparttable}
  }
\end{table*}

\subsection{Online Transfer Forecasting Results}
% Contrastingly, transfer forecasting necessitates training a model in one environment and subsequently testing it in another, thereby requiring the model to possess a heightened adaptability to unfamiliar data. We evaluate numerous forms of transfer forecasting, including intra-domain transfers (FaaS to FaaS, IaaS to IaaS), inter-domain transfers (FaaS to IaaS, IaaS to FaaS), and transfers across varying data granularities (ETTh1 to ETTm1, ETTm2 to ETTh2). Table \ref{tab:transfer_results} presents the cumulative performance. Overall, online models surpass the performance of retrained offline models in transfer tasks, suggesting that online predictions hold considerable advantages over offline predictions in cold-start scenarios when encountering new data. Specifically, the performance of our proposed \modelEGD far exceeded that of the baselines, which reduces MSE by 11.3\%, MAE by 12.7\%, and WMAPE by 13.3\% on average compared to the best baseline, OneNet , across six transfer tasks (\modelFTPL exceeded it by 3.6\%/7.7\%/6.1\%).
Online transfer forecasting involves training a model within one dateset and then applying it to another dataset, necessitating a substantially higher adaptability to unknown data.  In the transfer experiments, we evaluate the model's ability from multiple aspects:
\begin{enumerate}
    \item Transfer between datasets with the same resolution within the same domain (FaaS$\rightarrow$FaaS and IaaS$\rightarrow$IaaS).
    \item Transfer between datasets with the same resolution across different domains (FaaS$\rightarrow$IaaS and IaaS$\rightarrow$FaaS).
    \item Transfer between datasets with the different resolutions within the same domain (ETTh1$\rightarrow$ETTm1 and ETTm1$\rightarrow$ETTh2).
    \item Transfer between datasets with the different resolutions across different domains (FaaS$\rightarrow$Weather).
\end{enumerate}
The aggregate outcomes are detailed in Table \ref{tab:transfer_results}. Predominantly, online models outshine retrained offline models in transfer tasks, illustrating that online predictions possess significant advantages over offline predictions in scenarios characterized by cold starts and the introduction of new data. Notably, the \modelEGD's performance significantly surpasses that of established baselines, achieving a reduction in MSE by 15.3\%, MAE by 14.1\%, and WMAPE by 26.3\% on average, when compared to the leading baseline, OneNet, across a suite of six transfer tasks (with \modelFTPL also outperforming it, albeit by smaller margins of 3.1\%, 7.9\%, and 13.1\%).

\subsection{Ablation Study and Further Discussion}

\subsubsection{Ablation study}
% We analyzes the contribution of each \model's component. First, we explore the benefits of using the milti-input and multi-output mechanism (MIMO) by use a single patch size $16$. Second, we remove the online adaptor in the online Transformer backbone, which makes the backbone a naive transformer block. Specifically, by removing MIMO and online adaptor from \modelFTPL, it degrades into a PatchTST. The relevant results are displayed in the Table.\ref{tab:ablation} (the more comprehensive presentation, including results compared with PatchTST, is provided in the Table.\ref{tab:full_ablation_results} in Appendix. An important observation is that both components are very crucial, and losing either one will significantly degrade the performance. For \modelEGD, the MSE/MAE/WMAPE will increase 12.6\%/7.2\%/9.1\% without MIMO, and that will increase 29.4\%/14.3\%/13.5\% without online adaptor. The results are similar for \modelFTPL. Another possible observation is that in transfer tasks, online adaptor is especially important, without which MSE/MAE/WMAPE of \modelEGD will increase 54.2\%/25.4\%/21.0\% which illustrates the benefits of the online adaptor in helping the model address issues related to cold starts and domain drifts. Overall, these results demonstrates the complementary of each \model’s components to deal with different types of concept drift in time series.

Our analysis delves into the contribution of each component within the model. Initially, we investigate the advantages of employing the MIMO mechanism by utilizing a single patch size, without which \model consists of only one subnetwork, no longer forming an ensemble model. Subsequently, we omit the online adaptor in the online Transformer backbone, reducing it to a basic transformer block. Specifically, the removal of MIMO and the online adaptor from our model results in its simplification to a PatchTST. 
% \xiao{We need to clear state the relation between MIMO and online ensemble module.} 

\begin{table*}[!htbp]
  \caption{Results for the ablation study. We compare extensive competitive models under different prediction lengths. The input sequence length is set to 1440. \emph{Avg} is averaged from all four prediction lengths.}\label{tab:ablation}%\label{tab:full_ablation_results}
  \vskip 0.05in
  \centering
  % \resizebox{\linewidth}{!}
  {
  \begin{threeparttable}
  \begin{small}
  \renewcommand{\multirowsetup}{\centering}
  \setlength{\tabcolsep}{1pt}
  \begin{tabular}{c|c|>{\columncolor{light-gray}}c>{\columncolor{light-gray}}c>{\columncolor{light-gray}}c|ccc|ccc|>{\columncolor{light-gray}}c>{\columncolor{light-gray}}c>{\columncolor{light-gray}}c|ccc|ccc|ccc}
    \toprule
    \multicolumn{2}{c}{{Models}} & 
    \multicolumn{3}{c}{\rotatebox{0}{\cellcolor{light-gray}{{\modelEGD}}}} &
    \multicolumn{3}{c}{\rotatebox{0}{{{w/o MIMO}}}} &
    \multicolumn{3}{c}{\rotatebox{0}{{{w/o Online Adaptor}}}} &
    \multicolumn{3}{|c}{\rotatebox{0}{\cellcolor{light-gray}{\modelFTPL}}} &
    \multicolumn{3}{c}{\rotatebox{0}{{w/o MIMO}}} &
    \multicolumn{3}{c}{\rotatebox{0}{{w/o Online Adaptor}}} &
    \multicolumn{3}{c}{\rotatebox{0}{{Online PatchTST}}}\\
    \cmidrule(lr){3-5} \cmidrule(lr){6-8}\cmidrule(lr){9-11} \cmidrule(lr){12-14}\cmidrule(lr){15-17}\cmidrule(lr){18-20}\cmidrule(lr){21-23}
    \multicolumn{2}{c}{Metric} & \scalebox{0.8}{MSE} & \scalebox{0.8}{MAE} & \scalebox{0.8}{WMAPE} & 
    \scalebox{0.8}{MSE} & \scalebox{0.8}{MAE} & \scalebox{0.8}{WMAPE} & 
    \scalebox{0.8}{MSE} & \scalebox{0.8}{MAE} & \scalebox{0.8}{WMAPE} & 
    \scalebox{0.8}{MSE} & \scalebox{0.8}{MAE} & \scalebox{0.8}{WMAPE} & 
    \scalebox{0.8}{MSE} & \scalebox{0.8}{MAE} & \scalebox{0.8}{WMAPE} & 
    \scalebox{0.8}{MSE} & \scalebox{0.8}{MAE} & \scalebox{0.8}{WMAPE} & 
    \scalebox{0.8}{MSE} & \scalebox{0.8}{MAE} & \scalebox{0.8}{WMAPE}\\
    \toprule
    \multirow{5}{*}{\rotatebox{90}{\scalebox{1}FaaS}}
    &  \scalebox{1}{1} &\scalebox{1}{0.087} &\scalebox{1}{0.177}&\scalebox{1}{0.707}&\scalebox{1}{0.087 } &\scalebox{1}{0.198 }&\scalebox{1}{0.676 }&\scalebox{1}{0.094 } &\scalebox{1}{0.188 }&\scalebox{1}{0.765 }&\scalebox{1}{0.089} &\scalebox{1}{0.177}&\scalebox{1}{0.703}&\scalebox{1}{0.088 } &\scalebox{1}{0.194 }&\scalebox{1}{0.678 }&\scalebox{1}{0.092 } &\scalebox{1}{0.180 }&\scalebox{1}{0.698 }&\scalebox{1}{0.091} &\scalebox{1}{0.169}&\scalebox{1}{0.684}
    \\
    &  \scalebox{1}{10} &\scalebox{1}{0.174 } &\scalebox{1}{0.260 }&\scalebox{1}{1.240 }&\scalebox{1}{0.198 } &\scalebox{1}{0.280 }&\scalebox{1}{1.234 }&\scalebox{1}{0.184 } &\scalebox{1}{0.274 }&\scalebox{1}{1.378 }&\scalebox{1}{0.184} &\scalebox{1}{0.271}&\scalebox{1}{1.365}&\scalebox{1}{0.194 } &\scalebox{1}{0.276 }&\scalebox{1}{1.206 }&\scalebox{1}{0.188 } &\scalebox{1}{0.276 }&\scalebox{1}{1.487 }&\scalebox{1}{0.208} &\scalebox{1}{0.292}&\scalebox{1}{1.357}
    \\
    &  \scalebox{1}{30} &\scalebox{1}{0.217 } &\scalebox{1}{0.302 }&\scalebox{1}{1.487 }&\scalebox{1}{0.248 } &\scalebox{1}{0.336 }&\scalebox{1}{1.665 }&\scalebox{1}{0.223 } &\scalebox{1}{0.310 }&\scalebox{1}{1.632 }&\scalebox{1}{0.218} &\scalebox{1}{0.304}&\scalebox{1}{1.526}&\scalebox{1}{0.254 } &\scalebox{1}{0.340 }&\scalebox{1}{1.656 }&\scalebox{1}{0.230 } &\scalebox{1}{0.319 }&\scalebox{1}{1.730 }&\scalebox{1}{0.299} &\scalebox{1}{0.363}&\scalebox{1}{1.928}
    \\
    &  \scalebox{1}{60} &\scalebox{1}{0.354 } &\scalebox{1}{0.328 }&\scalebox{1}{1.688 }&\scalebox{1}{0.368 } &\scalebox{1}{0.340 }&\scalebox{1}{1.788 }&\scalebox{1}{0.362 } &\scalebox{1}{0.338 }&\scalebox{1}{1.829 }&\scalebox{1}{0.366} &\scalebox{1}{0.343}&\scalebox{1}{1.931}&\scalebox{1}{0.372 } &\scalebox{1}{0.348 }&\scalebox{1}{1.821 }&\scalebox{1}{0.369 } &\scalebox{1}{0.344 }&\scalebox{1}{1.961 }&\scalebox{1}{0.430} &\scalebox{1}{0.398}&\scalebox{1}{2.166}
    \\
    &  \scalebox{1}{\emph{Avg}} &\scalebox{1}{0.208 } &\scalebox{1}{0.267 }&\scalebox{1}{1.281 }&\scalebox{1}{0.225 } &\scalebox{1}{0.289 }&\scalebox{1}{1.341 }&\scalebox{1}{0.216 } &\scalebox{1}{0.278 }&\scalebox{1}{1.401 }&\scalebox{1}{0.214 } &\scalebox{1}{0.274 }&\scalebox{1}{1.381 }&\scalebox{1}{0.227 } &\scalebox{1}{0.290 }&\scalebox{1}{1.338 }&\scalebox{1}{0.220 } &\scalebox{1}{0.280 }&\scalebox{1}{1.469 }&\scalebox{1}{0.257 } &\scalebox{1}{0.306 }&\scalebox{1}{1.534 }
    \\
    \midrule
    \multirow{5}{*}{\rotatebox{90}{\scalebox{1}IaaS}}
    &  \scalebox{1}{1} &\scalebox{1}{0.421  } &\scalebox{1}{0.392  }&\scalebox{1}{0.474  }&\scalebox{1}{0.453 } &\scalebox{1}{0.402 }&\scalebox{1}{0.488 }&\scalebox{1}{0.481 } &\scalebox{1}{0.420 }&\scalebox{1}{0.504 }&\scalebox{1}{0.423  } &\scalebox{1}{0.388  }&\scalebox{1}{0.471  }&\scalebox{1}{0.457 } &\scalebox{1}{0.402 }&\scalebox{1}{0.483 }&\scalebox{1}{0.464 } &\scalebox{1}{0.411 }&\scalebox{1}{0.485 }&\scalebox{1}{0.476 } &\scalebox{1}{0.407 }&\scalebox{1}{0.494 }
    \\
    &  \scalebox{1}{10} &\scalebox{1}{0.552  } &\scalebox{1}{0.520  }&\scalebox{1}{0.622  }&\scalebox{1}{0.587 } &\scalebox{1}{0.533 }&\scalebox{1}{0.664 }&\scalebox{1}{0.584 } &\scalebox{1}{0.525 }&\scalebox{1}{0.629 }&\scalebox{1}{0.580 } &\scalebox{1}{0.524 }&\scalebox{1}{0.628 }&\scalebox{1}{0.587 } &\scalebox{1}{0.535 }&\scalebox{1}{0.666 }&\scalebox{1}{0.596 } &\scalebox{1}{0.533 }&\scalebox{1}{0.633 }&\scalebox{1}{0.611 } &\scalebox{1}{0.542 }&\scalebox{1}{0.665 }
    \\
    &  \scalebox{1}{30} &\scalebox{1}{0.735 } &\scalebox{1}{0.632 }&\scalebox{1}{0.734 }&\scalebox{1}{0.753 } &\scalebox{1}{0.641 }&\scalebox{1}{0.761 }&\scalebox{1}{0.756 } &\scalebox{1}{0.639 }&\scalebox{1}{0.745 }&\scalebox{1}{0.742 } &\scalebox{1}{0.641 }&\scalebox{1}{0.754 }&\scalebox{1}{0.754 } &\scalebox{1}{0.643 }&\scalebox{1}{0.766 }&\scalebox{1}{0.765 } &\scalebox{1}{0.653 }&\scalebox{1}{0.762 }&\scalebox{1}{0.792 } &\scalebox{1}{0.652 }&\scalebox{1}{0.777 }
    \\
    &  \scalebox{1}{60} &\scalebox{1}{0.828  } &\scalebox{1}{0.708  }&\scalebox{1}{0.829  }&\scalebox{1}{0.887 } &\scalebox{1}{0.747 }&\scalebox{1}{0.920 }&\scalebox{1}{0.856 } &\scalebox{1}{0.725 }&\scalebox{1}{0.848 }&\scalebox{1}{0.839 } &\scalebox{1}{0.706 }&\scalebox{1}{0.823 }&\scalebox{1}{0.885 } &\scalebox{1}{0.747 }&\scalebox{1}{0.915 }&\scalebox{1}{0.891 } &\scalebox{1}{0.761 }&\scalebox{1}{0.896 }&\scalebox{1}{0.876 } &\scalebox{1}{0.726 }&\scalebox{1}{0.867 }
    \\
    &  \scalebox{1}{\emph{Avg}} &\scalebox{1}{0.634 } &\scalebox{1}{0.563  }&\scalebox{1}{0.665  }&\scalebox{1}{0.670 } &\scalebox{1}{0.581 }&\scalebox{1}{0.708 }&\scalebox{1}{0.669 } &\scalebox{1}{0.577 }&\scalebox{1}{0.682 }&\scalebox{1}{0.646 } &\scalebox{1}{0.565  }&\scalebox{1}{0.669  }&\scalebox{1}{0.671 } &\scalebox{1}{0.582 }&\scalebox{1}{0.707 }&\scalebox{1}{0.679 } &\scalebox{1}{0.590 }&\scalebox{1}{0.694 }&\scalebox{1}{0.689 } &\scalebox{1}{0.582 }&\scalebox{1}{0.701 }
    \\
    \midrule
    \multirow{5}{*}{\rotatebox{90}{\scalebox{1}FaaS$\rightarrow$FaaS}}
    &  \scalebox{1}{1} &\scalebox{1}{0.161} &\scalebox{1}{0.185}&\scalebox{1}{0.028}&\scalebox{1}{0.184 } &\scalebox{1}{0.183 }&\scalebox{1}{0.027 }&\scalebox{1}{0.243 } &\scalebox{1}{0.208 }&\scalebox{1}{0.034 }&\scalebox{1}{0.235} &\scalebox{1}{0.185}&\scalebox{1}{0.027}&\scalebox{1}{0.182 } &\scalebox{1}{0.181 }&\scalebox{1}{0.026 }&\scalebox{1}{0.217 } &\scalebox{1}{0.218 }&\scalebox{1}{0.030 }&\scalebox{1}{0.374 } &\scalebox{1}{0.208 }&\scalebox{1}{0.028 }
    \\
    &  \scalebox{1}{10} &\scalebox{1}{0.291} &\scalebox{1}{0.313}&\scalebox{1}{0.042}&\scalebox{1}{0.472 } &\scalebox{1}{0.407 }&\scalebox{1}{0.068 }&\scalebox{1}{0.461 } &\scalebox{1}{0.378 }&\scalebox{1}{0.046 }&\scalebox{1}{0.456} &\scalebox{1}{0.335}&\scalebox{1}{0.043}&\scalebox{1}{0.490 } &\scalebox{1}{0.424 }&\scalebox{1}{0.070 } &\scalebox{1}{0.521 }&\scalebox{1}{0.414 }&\scalebox{1}{0.056 }&\scalebox{1}{0.657 } &\scalebox{1}{0.526 }&\scalebox{1}{0.058 }
    \\
    &  \scalebox{1}{30} &\scalebox{1}{0.459} &\scalebox{1}{0.371}&\scalebox{1}{0.043}&\scalebox{1}{0.500 } &\scalebox{1}{0.404 }&\scalebox{1}{0.068 }&\scalebox{1}{0.813 } &\scalebox{1}{0.580 }&\scalebox{1}{0.070 }&\scalebox{1}{0.482} &\scalebox{1}{0.419}&\scalebox{1}{0.051}&\scalebox{1}{0.499 } &\scalebox{1}{0.401 }&\scalebox{1}{0.061 }&\scalebox{1}{0.778 } &\scalebox{1}{0.570 }&\scalebox{1}{0.071 }&\scalebox{1}{1.157 } &\scalebox{1}{0.705 }&\scalebox{1}{0.078 }
    \\
    &  \scalebox{1}{60} &\scalebox{1}{0.672} &\scalebox{1}{0.452}&\scalebox{1}{0.055}&\scalebox{1}{0.709 } &\scalebox{1}{0.451 }&\scalebox{1}{0.064 }&\scalebox{1}{1.109 } &\scalebox{1}{0.606 }&\scalebox{1}{0.066 }&\scalebox{1}{0.824} &\scalebox{1}{0.439}&\scalebox{1}{0.057}&\scalebox{1}{0.704 } &\scalebox{1}{0.451 }&\scalebox{1}{0.064 }&\scalebox{1}{1.180 } &\scalebox{1}{0.673 }&\scalebox{1}{0.081 }&\scalebox{1}{1.276} &\scalebox{1}{0.715}&\scalebox{1}{0.084}
    \\
    &  \scalebox{1}{\emph{Avg}} &\scalebox{1}{0.396 } &\scalebox{1}{0.330 }&\scalebox{1}{0.042 }&\scalebox{1}{0.466 } &\scalebox{1}{0.361 }&\scalebox{1}{0.055 }&\scalebox{1}{0.657 } &\scalebox{1}{0.443 }&\scalebox{1}{0.054 }&\scalebox{1}{0.499 } &\scalebox{1}{0.345 }&\scalebox{1}{0.045 }&\scalebox{1}{0.469 } &\scalebox{1}{0.364 }&\scalebox{1}{0.055 }&\scalebox{1}{0.674 } &\scalebox{1}{0.469 }&\scalebox{1}{0.060 }&\scalebox{1}{0.866 } &\scalebox{1}{0.539 }&\scalebox{1}{0.062 }
    \\
    \midrule
    \multirow{5}{*}{\rotatebox{90}{\scalebox{0.8}{ETTh1$\rightarrow$ETTm1}}}
    &  \scalebox{1}{1} &{\scalebox{1}{0.088 }} &{\scalebox{1}{0.199 }} &{\scalebox{1}{0.096 }}&{\scalebox{1}{0.089 }} &{\scalebox{1}{0.199 }} &{\scalebox{1}{0.096 }}&{\scalebox{1}{0.091 }} &{\scalebox{1}{0.202 }} &{\scalebox{1}{0.099 }}&{\scalebox{1}{0.085 }} &{\scalebox{1}{0.200 }} &{\scalebox{1}{0.099 }}&{\scalebox{1}{0.086 }} &{\scalebox{1}{0.199 }} &{\scalebox{1}{0.096 }}&{\scalebox{1}{0.094 }} &{\scalebox{1}{0.205 }} &{\scalebox{1}{0.101 }}&{\scalebox{1}{0.092 }} &{\scalebox{1}{0.206 }} &{\scalebox{1}{0.101 }} 
    \\
    &  \scalebox{1}{10} &{\scalebox{1}{0.273 }} &{\scalebox{1}{0.366 }} &{\scalebox{1}{0.189 }}&{\scalebox{1}{0.366 }} &{\scalebox{1}{0.415 }} &{\scalebox{1}{0.208 }}&{\scalebox{1}{0.355 }} &{\scalebox{1}{0.407 }} &{\scalebox{1}{0.205 }}&{\scalebox{1}{0.354 }} &{\scalebox{1}{0.403 }} &{\scalebox{1}{0.198 }}&{\scalebox{1}{0.372 }} &{\scalebox{1}{0.420 }} &{\scalebox{1}{0.209 }}&{\scalebox{1}{0.364 }} &{\scalebox{1}{0.409 }} &{\scalebox{1}{0.206 }}&{\scalebox{1}{0.431 }} &{\scalebox{1}{0.453 }} &{\scalebox{1}{0.225 }} 
    \\
    &  \scalebox{1}{30} &{\scalebox{1}{0.466 }} &{\scalebox{1}{0.495 }} &{\scalebox{1}{0.269 }}&{\scalebox{1}{0.623 }} &{\scalebox{1}{0.574 }} &{\scalebox{1}{0.320 }}&{\scalebox{1}{0.659 }} &{\scalebox{1}{0.583 }} &{\scalebox{1}{0.303 }}&{\scalebox{1}{0.477 }} &{\scalebox{1}{0.501 }} &{\scalebox{1}{0.272 }}&{\scalebox{1}{0.611 }} &{\scalebox{1}{0.567 }} &{\scalebox{1}{0.315 }}&{\scalebox{1}{0.690 }} &{\scalebox{1}{0.602 }} &{\scalebox{1}{0.319 }}&{\scalebox{1}{0.770 }} &{\scalebox{1}{0.633 }} &{\scalebox{1}{0.328 }} 
    \\
    &  \scalebox{1}{60} &{\scalebox{1}{0.596 }} &{\scalebox{1}{0.559 }} &{\scalebox{1}{0.304 }}&{\scalebox{1}{0.746 }} &{\scalebox{1}{0.624 }} &{\scalebox{1}{0.342 }}&{\scalebox{1}{0.923 }} &{\scalebox{1}{0.697 }} &{\scalebox{1}{0.367 }}&{\scalebox{1}{0.640 }} &{\scalebox{1}{0.570 }} &{\scalebox{1}{0.317 }}&{\scalebox{1}{0.726 }} &{\scalebox{1}{0.620 }} &{\scalebox{1}{0.340 }}&{\scalebox{1}{0.910 }} &{\scalebox{1}{0.696 }} &{\scalebox{1}{0.369 }}&{\scalebox{1}{1.018 }} &{\scalebox{1}{0.732 }} &{\scalebox{1}{0.376 }} 
    \\
    &  \scalebox{1}{\emph{Avg}} &{\scalebox{1}{0.356 }} &{\scalebox{1}{0.405 }} &{\scalebox{1}{0.215 }}&{\scalebox{1}{0.456 }} &{\scalebox{1}{0.453 }} &{\scalebox{1}{0.241 }}&{\scalebox{1}{0.507 }} &{\scalebox{1}{0.472 }} &{\scalebox{1}{0.244 }}&{\scalebox{1}{0.389 }} &{\scalebox{1}{0.419 }} &{\scalebox{1}{0.222 }}&{\scalebox{1}{0.449 }} &{\scalebox{1}{0.452 }} &{\scalebox{1}{0.240 }}&{\scalebox{1}{0.515 }} &{\scalebox{1}{0.478 }} &{\scalebox{1}{0.249 }}&{\scalebox{1}{0.578 }} &{\scalebox{1}{0.506 }} &{\scalebox{1}{0.258 }} 
    \\
    \bottomrule
  \end{tabular}
  %   \begin{tablenotes}
  %       \footnotesize
  %       \item[] $\ast$ means that there are some mismatches between our input-output setting and their papers. We adopt their official codes and only change the length of input and output sequences for a fair comparison.
  % \end{tablenotes}
    \end{small}
  \end{threeparttable}
  }
\end{table*}

These pertinent findings are recorded in Table \ref{tab:ablation}. A key observation is the critical nature of both components, where the absence of either leads to a substantial decline in performance. For instance, excluding MIMO from \modelEGD results in an increase in MSE/MAE/WMAPE by 15.0\%, 8.2\%, and 13.4\%, respectively, and the removal of the online adaptor sees these metrics escalate by 29.4\%, 14.3\%, and 13.5\%, respectively. These trends are consistent for \modelFTPL. Notably, in transfer tasks, the importance of the online adaptor is magnified, with its exclusion causing MSE/MAE/WMAPE in \modelEGD to soar by 54.2\%, 25.4\%, and 21.0\%, respectively, underscoring the adaptor's role in tackling cold starts and domain shifts. %Collectively, these outcomes highlight the complementary roles of the model's components in addressing various forms of concept drift within time series data.

\subsubsection{Efficiency analysis}

\begin{figure}[htbp]
    \centering
    % \hspace{-30mm}
    \begin{minipage}{0.46\textwidth}
    \includegraphics[width=1\linewidth]{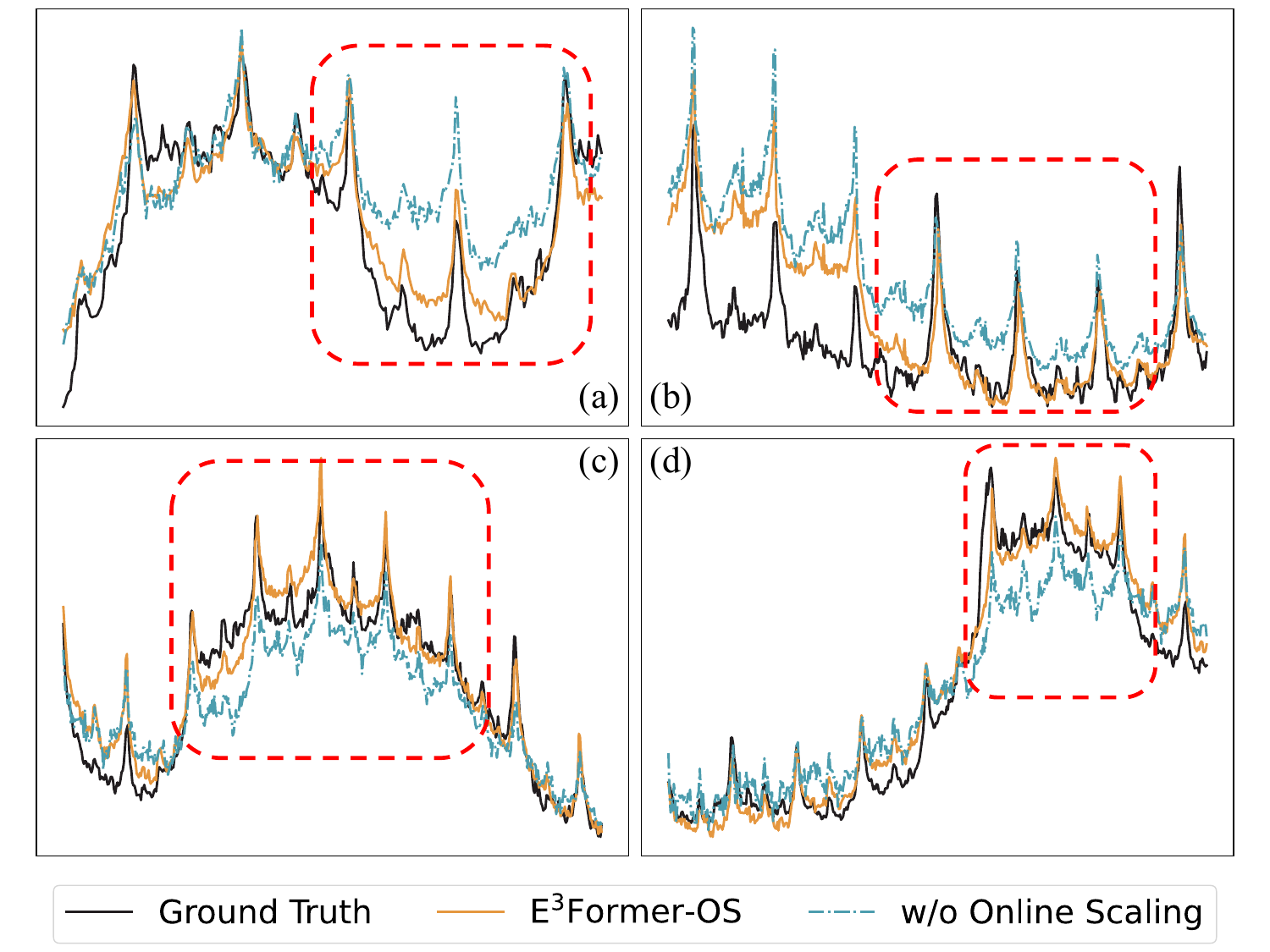}
    \vspace{-4mm}
    \caption{Visualizing \modelEGD’s forecasting.}
    \label{fig:forecasting_fix}
    \end{minipage}
\end{figure}

% We calculate the average online inference time for different online models to test their efficiency. Figure~\ref{fig:inference_time} demonstrates these results. The inference time of OneNet is significantly higher than that of other models because it serially employs two FSNet models. Compared with OneNet, the inference efficiency of \modelEGD increases by 30.8\% on average. Although \model incorporates more base models (four independent sub-networks), thanks to MIMO configurations and channel independence setting, the model exhibits good parallelism, which can significantly reduce inference time while generating more accurate predictions. Further, as the Follow-The-Perturbed-Leader algorithm's simplicity nature, it reaches a 39.6\% decrease of inference time than OneNet on average.

\begin{figure}[!htbp]
    \centering
    % \hspace{-30mm}
    \begin{minipage}{0.46\textwidth}
    \includegraphics[width=\linewidth]{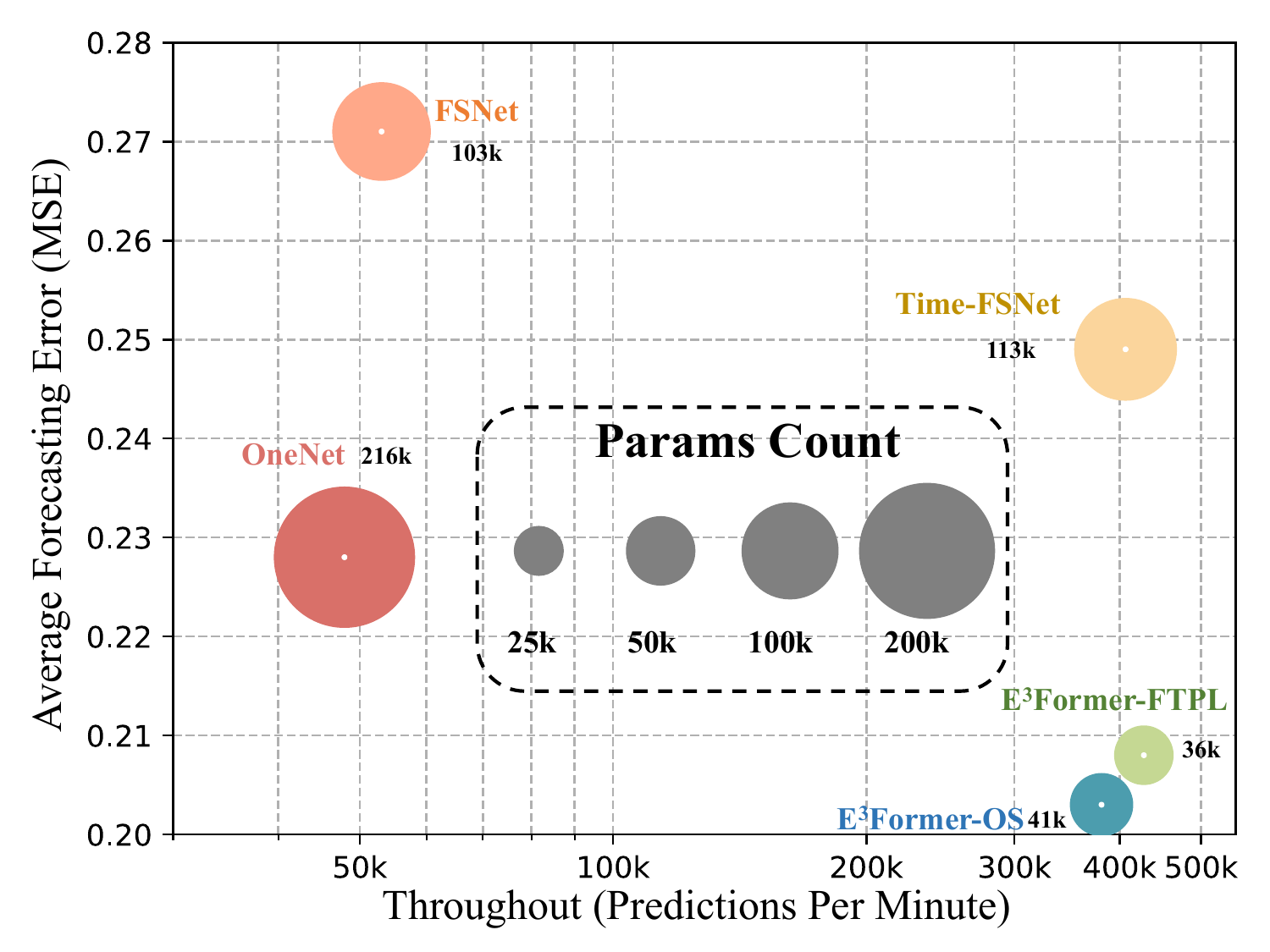}
    \vspace{-4mm}
    \caption{Model performance and throughout comparison.}
    % \vspace{-1mm}
    \label{fig:throughout}
    \vspace{-2mm}
    \end{minipage}
\end{figure}

%\xiao{Figure 3 is not described. Maybe we can keep Figure 3 and leave out Figure 5(a).} 
We evaluated the parameter count and the average throughput of various online models to measure their efficiency, with results illustrated in Fig.~\ref{fig:throughout}. Each model was tested using Nvidia Tesla V100 32GB GPUs. Among these, \modelEGD and \modelFTPL stand out for delivering not only precise forecasts but also the highest throughput levels. OneNet's inference time was notably longer than that of other models. This increased time is attributed to OneNet's ensemble setup, which involves the sequential application of two FSNet models. \modelFTPL showed a significant reduction in inference time—39.6\% less than that of OneNet. Additionally, the parameter count of \model is significantly reduced, for example, \modelFTPL constituting merely 16.7\% of OneNet's total. This reduction enables it to attain a throughput that astonishingly exceeds OneNet's by over \textbf{800\%}.

%The inference time for OneNet is markedly higher compared to other models due to its unique configuration, which sequentially applies two FSNet models. Relative to OneNet, the inference efficiency of the \modelEGD model is enhanced by an average of 30.8\%. Despite \model incorporating a larger number of base models (comprising four independent sub-networks), its implementation of MIMO configurations and the setting for channel independence significantly bolster parallelism. This approach not only reduces inference time but also ensures more precise predictions. Moreover, leveraging the inherently simplistic design of the Follow-The-Perturbed-Leader algorithm, it achieves a reduction in inference time by 39.6\% on average compared to OneNet.

\subsubsection{Forecasting results visualization} 
\begin{figure}[htbp]
    \centering
    \hspace{60mm}
    \begin{minipage}{0.46\textwidth}
    \vspace{-4mm}
    \includegraphics[width=\linewidth]{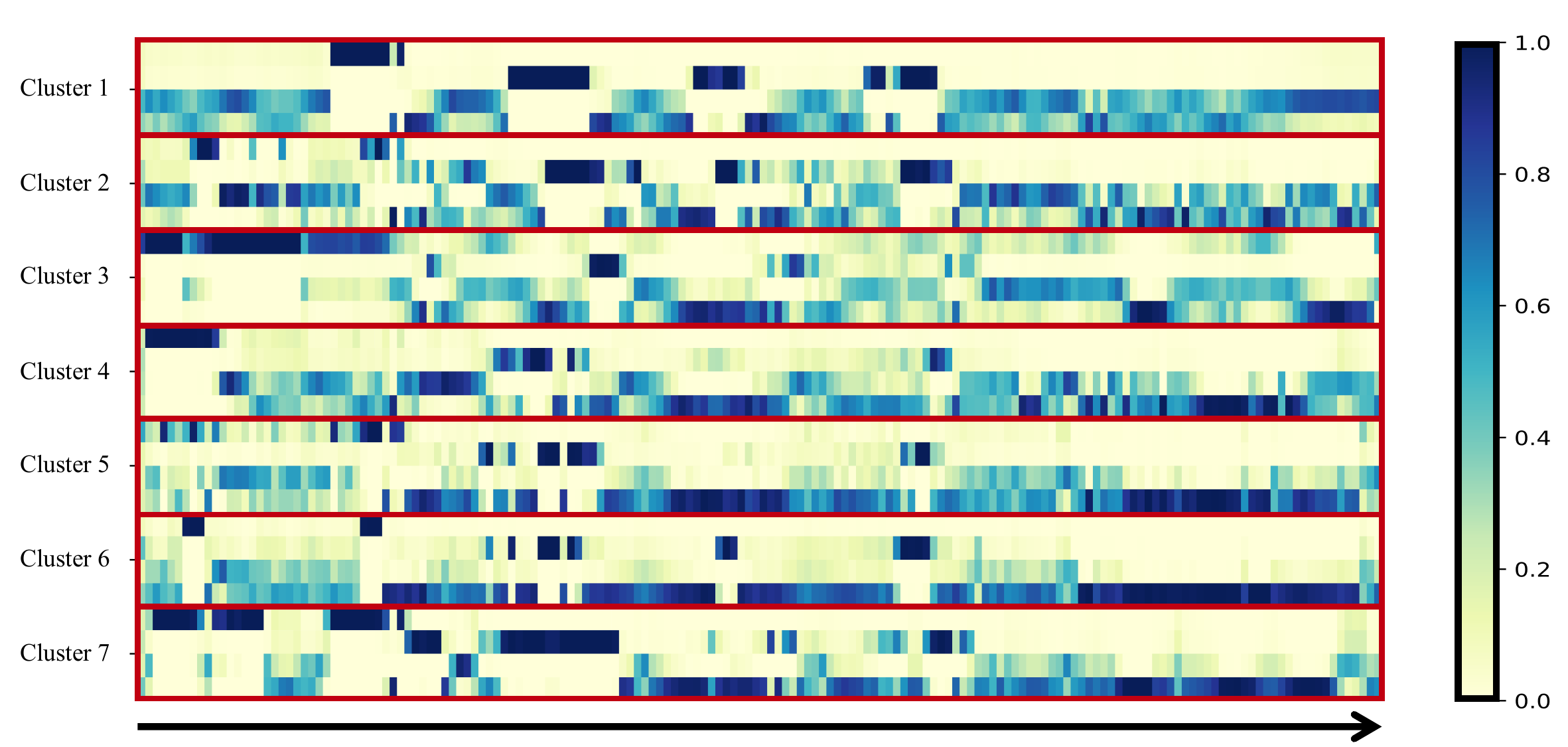}
    \vspace{-4mm}
    \caption{The variation of weights among different base models during the online inference process of \modelEGD.}
    \label{fig:weights}
    \vspace{-2mm}
    \end{minipage}
\end{figure}

Some examples of forecasting results for \modelEGD on FaaS are visualized in Fig.~\ref{fig:forecasting_fix}. The dashed cyan line represents the prediction results of \modelEGD without Online Scaling. We can intuitively observe that after applying Online Scaling, the prediction results better fit the real sequence, especially during the process of trend drifts in the sequence. Additionally, the illustration in Fig.~\ref{fig:weights} showcasing the dynamic changes in the weights of different base models within \modelEGD during the inference process, underscores the model's ability to swiftly adapt to shifts in sequential patterns. This rapid adjustment 
of weights is a crucial aspect that addresses the issues "slow switch phenomenon"~\cite{cesa2006prediction}, 
where static or slow-adapting models often struggled to maintain high predictive accuracy in the face of evolving data distributions.

\subsubsection{Effects of number of subnetworks}
\begin{figure}[htbp]
    \centering
    \hspace{60mm}
    \begin{minipage}{0.46\textwidth}
    \includegraphics[width=\linewidth]{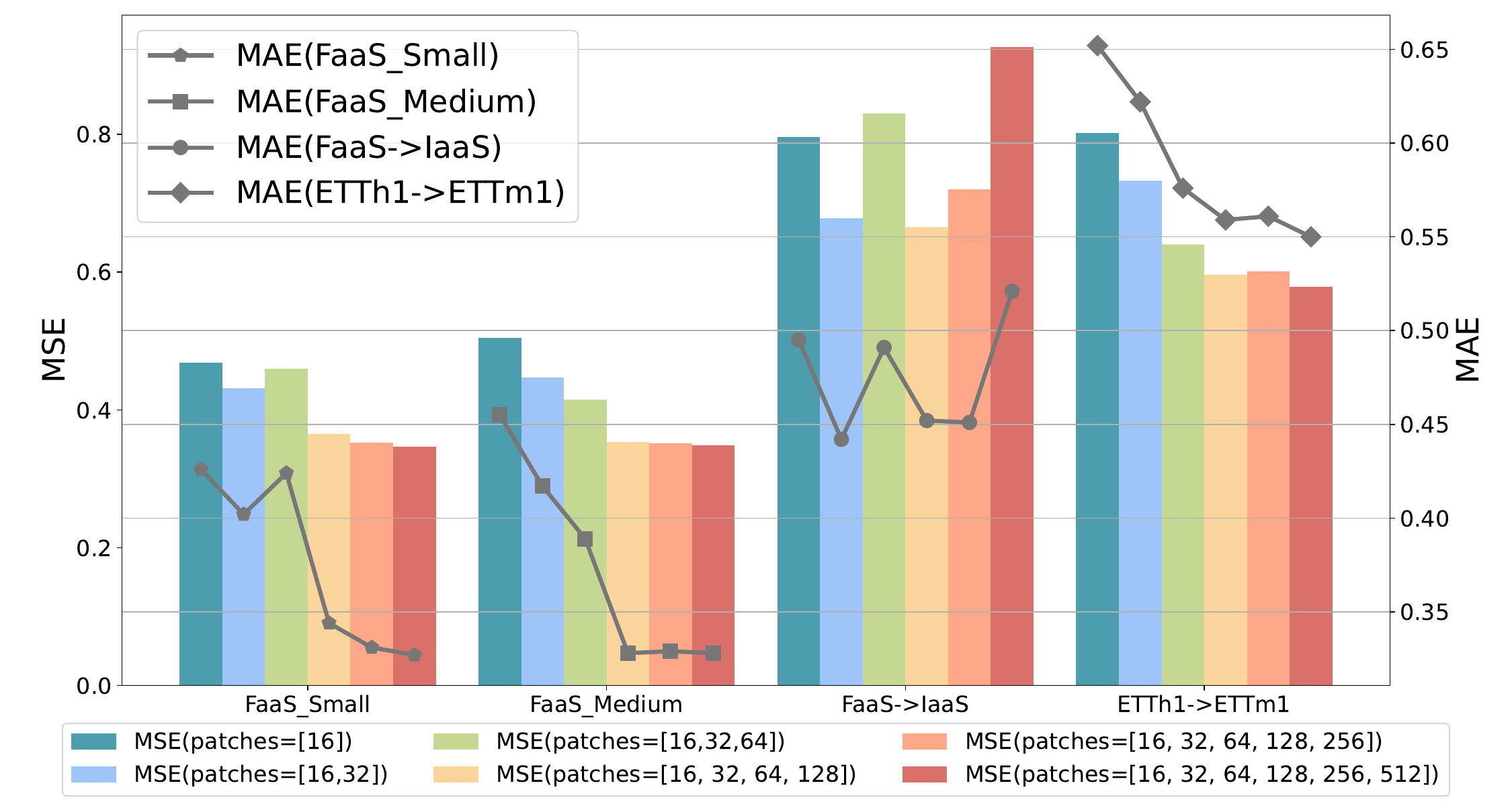}
    \vspace{-4mm}
    \caption{The performance of \modelEGD as the number of subnetworks varies from $1$ to $6$.}
    \label{fig:number}
    \vspace{-2mm}
    \end{minipage}
\end{figure}

Fig.~\ref{fig:number} shows the performance of \modelEGD as the patch size group varies. \model with a patch size group $\{16\}$ is equivalent to a PatchTST since there is only one subnetwork. As the grows of the number of subnetworks, we can see that the errors of \modelEGD slowly decline as they utilize more of the network capacity. The number of subnetworks contained in patch sizes $\{16,32,64,128\}$ and $\{16, 32, 64, 128, 256, 512\}$ differs by 50\%, yet their performance is comparable, with the former even outperforming the latter in some cases (FaaS$\rightarrow$IaaS). This could be attributed to the fact that the network capacity is nearing its limit, whereafter the improvement in model performance due to an increase in the number of subnetworks exhibits a diminishing marginal effect. Therefore, considering the balance between efficiency and effectiveness, we generally utilized models with 4 subnetworks (patch sizes $\{16, 32, 64, 128\}$) in the experiments.%, even though their performance might be slightly inferior to that of models with more subnetworks.

\subsection{Kubernetes Horizontal Pod Autoscaling}
To explore the value of workload forecasting models in real-world tasks, the best approach would be to deploy them in online auto-scaling systems. However, online environments are often highly complex, comprising numerous components and rules, and the complexity makes it infeasible to isolate and measure the performance of forecasting models using controlled experiments. Moreover, online services typically have very low tolerance for high latency, making it difficult to conduct long-term online tests at the risk of degrading service quality. These factors render it impractical to rely on online testing to assess the value of forecasting models.

\begin{figure}[!tbp]
    \centering
    \hspace{60mm}
    \begin{minipage}{0.46\textwidth}
    \includegraphics[width=\linewidth]{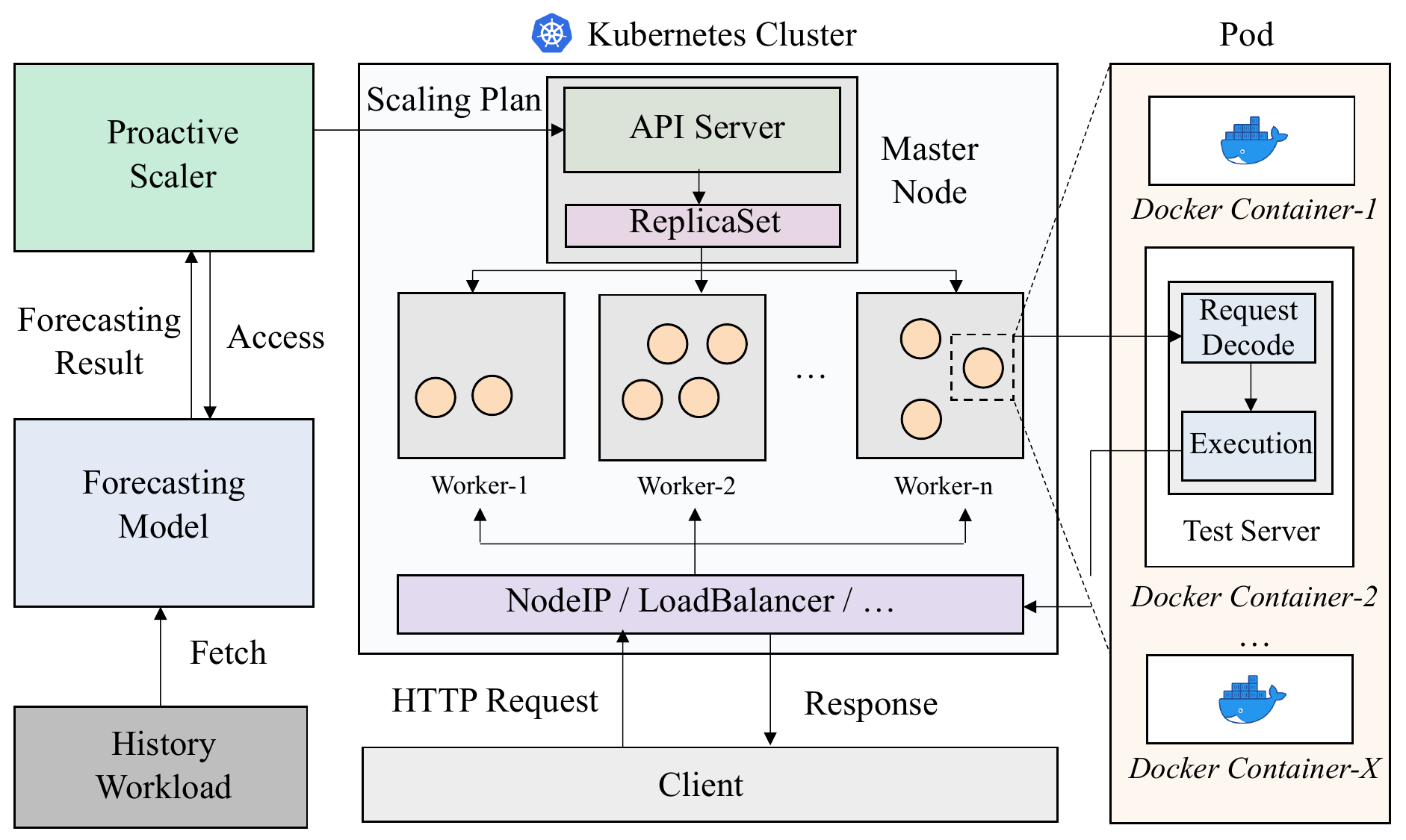}
    % \vspace{-4mm}
    \caption{kubernetes Horizontal Pod Auto-Scaler System.}
    \label{fig:k8s}
    \vspace{-4mm}
    \end{minipage}
\end{figure}

To evaluate the performance of \model in a real-world production setting, we construct a controlled Kubernetes system and perform a comparative analysis against other forecasting models within the framework of Kubernetes Horizontal Pod Autoscaling (HPA). Prior research, including AHPA\cite{zhou2023ahpa}, has highlighted the benefits of predictive HPA over traditional reactive approaches, attributing these advantages to the mitigation of performance degradation due to pod initialization delays. In our experiment, as shown in Figure~\ref{fig:k8s}, we deploy a test service on a Kubernetes cluster, subjecting it to a workload pattern derived from historical QPS data sourced from a FaaS cluster. 
We conduct offline model training based on the data collected by the system over the past two weeks.
At the commencement of each scaling interval (defaulting to 1 minute), scaling decisions are executed based on the predicted workload for that interval.
At regular intervals (defaulting to 10 minutes), the system review forecasting errors based on real-time metrics and performs online model updates.

We integrate each forecasting model into the Kubernetes HPA system and monitor the latency and resource consumption of the pods over a two-week period. For the Ideal strategy, these decisions are informed by actual future workload data. Our analysis encompasses comparisons between HPAs leveraging various forecasting models, the Na\"ive HPA included with Kubernetes, and the Ideal HPA with foresight into future workloads (AHPA is excluded as it is not open-sourced), with results presented in Table \ref{tab:hpa}. It is important to recognize that assessing HPA based solely on latency or resource utilization is not appropriate. If we were to allocate nearly unlimited physical resources to the system, latency would be minimal (or, conversely, if we allocated only one Pod, resource usage would be minimal, though latency would be uncontrollable). However, this is not aligned with best practices for cloud systems. 

\begin{table}[!htp]
  \caption{Results of predictive Kubernetes Horizontal Pod Autoscaling. \textit{Ave} stands for Average, \textit{99.9-Lat}, \textit{99-Lat} and \textit{90-Lat} represents the 99.9, 99 and 90 percentile latency. }
  \vspace{-2mm}
  \label{tab:hpa}
  \vskip 0.05in
  \centering
  \resizebox{\linewidth}{!}{
  \begin{threeparttable}
  % \begin{small}
  \renewcommand{\multirowsetup}{\centering}
  \setlength{\tabcolsep}{1pt}
  \begin{tabular}{c|c|c|c|c|c|c|c}
    \toprule
    \scalebox{1}{Models} & \scalebox{1}{Ave-Lat(s)} & \scalebox{1}{Max-Lat(s)} & \scalebox{1}{99.9-Lat(s)} & \scalebox{1}{99-Lat(s)} & \scalebox{1}{90-Lat(s)} & \scalebox{1}{AvePod} & \scalebox{1}{MaxPod} \\
    \scalebox{1}{iTransformer} & \scalebox{1}{0.264}& \scalebox{1}{63.156} & \scalebox{1}{3.277}  & \scalebox{1}{1.224} & \scalebox{1}{0.422} & \scalebox{1}{15.610} & \scalebox{1}{19} \\
    \scalebox{1}{FSNet} & \scalebox{1}{0.236}& \scalebox{1}{7.347} & \scalebox{1}{2.321}  & \scalebox{1}{1.039} & \scalebox{1}{0.405} & \scalebox{1}{14.942} & \scalebox{1}{29} \\
    \scalebox{1}{OneNet} & \scalebox{1}{0.267}& \scalebox{1}{52.906} & \scalebox{1}{3.085}  & \scalebox{1}{1.217} & \scalebox{1}{0.430} & \scalebox{1}{14.949} & \scalebox{1}{22} \\
    \midrule
    \scalebox{1}{Na\"ive HPA} & \scalebox{1}{0.231} & \scalebox{1}{91.022} & \scalebox{1}{8.275}  & \scalebox{1}{0.689} & \scalebox{1}{0.343} & \scalebox{1}{16.574} & \scalebox{1}{34} \\
    \midrule
    \scalebox{1}{Ideal} & \scalebox{1}{\textbf{0.219}}& \scalebox{1}{\textbf{9.419}} & \scalebox{1}{\textbf{2.087}}  & \scalebox{1}{\textbf{0.739}} & \scalebox{1}{\textbf{0.371}} & \scalebox{1}{\textbf{14.608}} & \scalebox{1}{\textbf{24}} \\
    \midrule
    \scalebox{1}{\modelFTPL} & \scalebox{1}{0.223}& \scalebox{1}{7.734} & \scalebox{1}{2.072}  & \scalebox{1}{0.767} & \scalebox{1}{0.380} & \scalebox{1}{15.071} & \scalebox{1}{22} \\
    \scalebox{1}{\modelEGD} & \scalebox{1}{\textbf{0.218}}& \scalebox{1}{\textbf{7.535}} & \scalebox{1}{\textbf{1.953}} & \scalebox{1}{\textbf{0.731}} & \scalebox{1}{\textbf{0.368}} & \scalebox{1}{\textbf{15.368}} & \scalebox{1}{\textbf{21}} \\
    \bottomrule
  \end{tabular}
  \end{threeparttable}
  }
\end{table}

Our research indicates that the \model effectively forecasts future workloads, thereby enhancing Quality of Service (QoS). This improvement is evidenced by a roughly 30\% reduction in the 99th percentile (p99) latency and efficient resource utilization, as reflected by the average number of pods. Comparatively, \model achieves lower latency and uses fewer pods than other forecasting models. For example, \modelFTPL demonstrates a 16.5\% improvement in average latency over OneNet, without a significant increase in pod usage. We note that HPAs guided by other forecasting models have also performed well in certain metrics. For instance, the iTransformer-based HPA strategy has the lowest maximum pod occupation, though with the highest average latency. Conversely, the FSNet-based HPA strategy has the lowest maximum latency (Max-Lat), but with the highest number of maximum pod occupations. In contrast, \model strikes the best balance between latency and resource utilization. Moreover, HPAs based on \model outperform the Na"ive HPA in both average and peak latency metrics while using fewer pods. Notably, \modelEGD reduces maximum latency by 91.7\% compared to the Na\"ive HPA. \model offers a similar QoS to the ideal HPA, albeit with a slightly higher average number of pods.

\begin{table}[!htp]
  \caption{Results of predictive Kubernetes Horizontal Pod Autoscaling in cold start (transfer forecasting) setting. The source FaaS cluster is selected randomly from history record, and the target cluster is the same as Table~\ref{tab:hpa}. }
  \vspace{-2mm}
  \label{tab:hpa2}
  \vskip 0.05in
  \centering
  \resizebox{\linewidth}{!}{
  \begin{threeparttable}
  % \begin{small}
  \renewcommand{\multirowsetup}{\centering}
  \setlength{\tabcolsep}{1pt}
  \begin{tabular}{c|c|c|c|c|c|c|c}
    \toprule
    \scalebox{1}{Models} & \scalebox{1}{Ave-Lat(s)} & \scalebox{1}{Max-Lat(s)} & \scalebox{1}{99.9-Lat(s)} & \scalebox{1}{99-Lat(s)} & \scalebox{1}{90-Lat(s)} & \scalebox{1}{AvePod} & \scalebox{1}{MaxPod} \\
    \scalebox{1}{iTransformer} & 0.334 & 97.692 & 9.331 & 1.977 & 0.577 & 16.291 & 27\\
    \scalebox{1}{FSNet} & 0.254 & 14.338 & 2.971 & 1.313 & 0.424 & 15.449 & 33 \\
    \scalebox{1}{OneNet} & 0.269 & 52.741 & 3.221 & 1.232 & 0.443 & 15.841 & 26 \\
    \midrule
    \scalebox{1}{Na\"ive HPA} & \scalebox{1}{0.231} & \scalebox{1}{91.022} & \scalebox{1}{8.275}  & \scalebox{1}{0.689} & \scalebox{1}{0.343} & \scalebox{1}{16.574} & \scalebox{1}{34} \\
    \midrule
    \scalebox{1}{Ideal} & \scalebox{1}{\textbf{0.219}}& \scalebox{1}{\textbf{9.419}} & \scalebox{1}{\textbf{2.087}}  & \scalebox{1}{\textbf{0.739}} & \scalebox{1}{\textbf{0.371}} & \scalebox{1}{\textbf{14.608}} & \scalebox{1}{\textbf{24}}   \\
    \midrule
    \scalebox{1}{\modelFTPL} & \textbf{0.223} & \textbf{12.317} & \textbf{2.311} & \textbf{0.771} & \textbf{0.379} & \textbf{15.229} & \textbf{26}  \\
    \scalebox{1}{\modelEGD} & 0.232 & 15.691 & 2.396 & 0.766 & 0.368 & 15.393 & 25\\
    \bottomrule
  \end{tabular}
  \end{threeparttable}
  }
\end{table}

Current predictive auto-scaling systems are bottlenecked by their reliance on historical data; for example, for predictive auto-scaling, AWS ECS\footnotemark[6] requires at least 24 hours (two weeks being ideal) of historical data (3, 7, 7 days for Google Cloud\footnotemark[7], Azura\footnotemark[8], and Alibaba Cloud\footnotemark[9], respectively), which restricts their deployment with limited historical data (so-called \emph{cold-start} tasks). To evaluate our model's performance in cold-start scenarios, we conducted kubernetes HPA experiments where the forecasting models were pretrained offline on data from other FaaS clusters and then fine-tuned online using data from the target cluster for cross-data transfer predictions, the results are shown in Table~\ref{tab:hpa2}. 

\footnotetext[6]{{\url{https://docs.aws.amazon.com/AmazonECS/latest/developerguide/predictive-auto-scaling.html}}}
\footnotetext[7]{{\url{https://cloud.google.com/compute/docs/autoscaler/predictive-autoscaling}}}
\footnotetext[8]{{\url{https://techcommunity.microsoft.com/blog/azureobservabilityblog/general-availability-predictive-autoscaling-for-vmss/3652844}}}
\footnotetext[9]{{\url{https://help.aliyun.com/zh/ack/serverless-kubernetes/user-guide/deploy-ahpa?spm=a2c4g.11186623.0.0.27197c8dzSnS17}}}

Due to the lack of historical data, the forecasting errors of the models is often larger at the beginning of online forecasting, which affects the effectiveness of auto-scaling (as evidenced by higher maximum latency and maximum Pod occupations compared to the results in Table~\ref{tab:hpa} for all models). However, thanks to the targeted design of \model (including the Adapter and Ensembler), it can adapt more quickly to the dynamic trends of the sequence and produce more robust forecastings. As a result, metrics such as average latency, 90th percentile (p90) latency, and average Pod occupations are comparable to those in Table~\ref{tab:hpa}. In particular, \modelFTPL achieves results closest to the Ideal HPA strategy, which can be attributed to the simplicity of the FTPL algorithm, which endows it with the better generalizability.

\section{Conclusion}

We design a novel ensemble learning model specifically for the nuanced demands of workload forecasting in large-scale cloud systems. 
Our model synergizes the predictive capabilities of multiple subnetworks based on a shared backbone, thus ensuring superior accuracy in online scenarios, especially in cold starts. 
Remarkably, it accomplishes this with a minimal increase in computational overhead, adhering to the lean operational ethos of serverless systems. 
To validate the effectiveness of the model, we have collected a workload dataset of large-scale cloud computing system and open-sourced it to advance community research.
Through extensive experimentation on the real-world datasets, we establish the efficacy of our ensemble model. 
Currently, our method has been deployed within ByteDance's IHPA platform, which supports the stable operation of over 30 applications, such as Douyin E-Commerce and Toutiao. The predictive auto-scaling capacity reaching over 600,000 CPU cores. On the basis of essentially ensuring service quality, the predictive auto-scaling system can reduce resource utilization by over 40\%.
In the future, we plan to explore more efficient and accurate predictive auto-scaling systems to support the efficient and lean operation of increasingly large-scale computing and storage tasks.
% The findings not only illustrate our model's enhanced predictive precision but also its adeptness in dynamically adapting to the ever-changing landscapes of serverless computing. 

% Although our motivation originates from the cloud computing application, the targeted challenges are general problems for time series forecasting. Thus, our model has potential to be extended to a broader area.

% The method has its limitations. The proposed model uses a data-hungry Transformer as the backbone model, therefore it requires more data in the pre-training phase to avoid overfitting compared to other online methods.
% In the future, we will explore the possibility of integrating the proposed online framework with recent Time Series Foundation Models~\cite{das2023decoder,liuitransformer} to mitigate this impact.
% Moreover, patch sizes in the proposed model significantly affect forecasting performance, especially for data with seasonality. Adjusting the model for data with unseen seasonality is not explored in this paper and will be a future research direction.

%\clearpage

\bibliographystyle{ACM-Reference-Format}
\bibliography{reference}

\end{document}